\documentclass[authoryear,preprint,12pt]{elsarticle}

\usepackage{amssymb}
\usepackage{amsmath}
\usepackage{amsthm}

\usepackage{lineno}

\usepackage{graphicx}%
\usepackage{multirow}%
\usepackage{amsmath,amssymb,amsfonts}%
\usepackage{amsthm}%
\usepackage{mathrsfs}%
\usepackage[title]{appendix}%
\usepackage{xcolor}%
\usepackage{textcomp}%
\usepackage{manyfoot}%
\usepackage{booktabs}%
\usepackage{algorithm}%
\usepackage{algorithmicx}%
\usepackage{algpseudocode}%
\usepackage{listings}%

\usepackage{algorithm}
\usepackage{algpseudocode}
\usepackage{glossaries}
\usepackage{comment}
\usepackage{annotate-equations}
\usepackage{soul}
\usepackage{subcaption}
\usepackage{multirow}
\usepackage{hhline}
\usepackage{adjustbox}
\usepackage{url}
\usepackage[colorlinks=true]{hyperref}

\makeglossaries
\newacronym{xai}{\textit{XAI}}{\textit{Explainable Artificial Intelligence}}
\newacronym{pg}{\textit{PG}}{\textit{Policy Graph}}
\newacronym{ma}{\textit{MA}}{\textit{Multi-Agent}}
\newacronym{rl}{\textit{RL}}{\textit{Reinforcement Learning}}
\newacronym{chr}{\textit{CHR}}{\textit{Causal History of Reasons}}
\newacronym{ef}{\textit{EF}}{\textit{Enabling factors}}
\newacronym{ppo}{\textit{PPO}}{\textit{Proximal Policy Optimisation}}
\glsaddall

\newcommand{\ie}{\textit{i}.\textit{e}.~}
\newcommand{\eg}{\textit{e}.\textit{g}.~}
\newcommand{\entropy}{\mathrm{H}}
\newcommand{\realstate}[1]{\mathtt{#1}}

\newcommand{\cmasone}{Human-Collaborating Agent} 
\newcommand{\marcone}{PPO Agent 1} 
\newcommand{\cmashuman}{Human Agent} 
\newcommand{\marctwo}{PPO Agent 2} 
\newcommand{\randomagent}{Random Agent} 

\newcommand{\simple}{Simple} 
\newcommand{\randomzero}{Random 0} 
\newcommand{\randomone}{Random 1} 
\newcommand{\randomthree}{Random 3} 
\newcommand{\unident}{Unident\_s} 

\journal{Expert Systems with Applications}

\begin{document}

\begin{frontmatter}



\title{Intention-aware policy graphs: answering \emph{what}, \emph{how}, and \emph{why} in opaque agents}

\author[bsc]{Victor Gimenez-Abalos}
\ead{victor.gimenez@bsc.es}
\author[upc,bsc]{Sergio Alvarez-Napagao\corref{cor1}}
\ead{salvarez@cs.upc.edu}
\cortext[cor1]{Corresponding author}
\author[bsc]{Adrian Tormos}
\ead{adrian.tormos@bsc.es}
\author[upc,bsc]{Ulises Cortés}
\ead{ulises.cortes@bsc.es}
\author[upc]{Javier Vázquez-Salceda}
\ead{jvazquez@cs.upc.edu}
\affiliation[bsc]{organization={Barcelona Supercomputing Center},
            addressline={Plaça Eusebi Guell, 1-3}, 
            city={Barcelona},
            postcode={08034}, 
            country={Spain}}
\affiliation[upc]{organization={Universitat Politecnica de Catalunya},
            addressline={c/Jordi Girona, 1-3}, 
            city={Barcelona},
            postcode={08034}, 
            country={Spain}}
            
\begin{abstract}
Agents are a special kind of AI-based software in that they interact in complex environments and have increased potential for emergent behaviour. 
Explaining such emergent behaviour is key to deploying trustworthy AI, but the increasing complexity and opaque nature of many agent implementations makes this hard. 
In this work, we propose a Probabilistic Graphical Model along with a pipeline for designing such model --by which the behaviour of an agent can be deliberated about-- and for computing a robust numerical value for the intentions the agent has at any moment. 
We contribute measurements that evaluate the interpretability and reliability of explanations provided, and enables explainability questions such as `what do you want to do now?' (\eg deliver soup) `how do you plan to do it?' (\eg returning a plan that considers its skills and the world), and `why would you take this action at this state?' (\eg explaining how that furthers or hinders its own goals). 
This model can be constructed by taking partial observations of the agent's actions and world states, and we provide an iterative workflow for increasing the proposed measurements through better design and/or pointing out irrational agent behaviour.
\end{abstract}


\end{frontmatter}

\section{Introduction}\label{sec:introduction}


Among the tasks within the purview of Artificial Intelligence (AI), the issue of solving problems without giving explicit knowledge on \textit{how\/} to solve them is very pervasive. 
However, precisely because of the definition of such a task, the result is an artefact that,  unless explicitly designed to be transparent, is often not interpretable or, hence, trustworthy~\citep{zhang_survey_2021,lipton_mythos_2017}. This is where the field of \gls{xai} shines through.

A model explanation is an exercise in communication between a sender or source (\ie the model or one of its components) and a receiver (\ie the explainee, a human or another processor for a downstream task) that describes the relevant context or the causes surrounding some facts~\citep{lewis_causal_1986,miller_explanation_2019,wright_explanation_2004}, which in the context of AI is often related to its final or intermediary outputs or decisions. 
Any such communicative act can be considered an explanation, but not all explanations may be useful or even desirable. 
According to empirical studies~\citep{slugoski_attribution_1993}, it can be argued that the form of an explanation must depend on its function as an answer to a question within a conversational framework.
Furthermore, in the words of Herbert Paul Grice~\citep{cole_logic_1975},
for a communicative act to be useful, four maxims should be followed: 

\begin{enumerate}
    \item \textbf{Manner}: the message or \textit{explanans\/} should be comprehensible and clear to the receiver, which within the context of \gls{xai} is often referred to as \textit{interpretability}~\citep{lipton_mythos_2017}, 
    \item \textbf{Quality}: the message contains truthful information; in the context of \gls{xai}, \textit{reliability} or explanation verification~\citep{zhou_feature_2021,slack_reliable_2021,arias-duart_focus_2022},
    \item \textbf{Quantity}: the length of a message should be just enough to be informative, often a heuristic implicitly agreed upon in the design of explainable systems which depends on both the sender and the code it uses, and
    \item \textbf{Relation}: the explanation should be relevant to the given context, significant when one can keep searching for causes of causes beyond the scope of relevance.
\end{enumerate}

Therefore, by following these maxims, we can identify specific metrics (interpretability, reliability, length, relevance) that allow us to measure specific interesting properties of the explanations and place them in a metric space that allows us to generate comparisons.
In this paper, we focus on the first two: reliability (\ie whether the explanation given by the model is factually correct and coherent over its behaviour, dependent solely on the sender); and interpretability (\ie whether the produced communicative act is something that the receiver can comprehend or use correctly, which is dependent on the receiver). 
These two metrics are separate optimisation objectives, which tend to be in conflict.
For instance, consider a complex machine learning model. The most reliable explanation would involve a detailed breakdown of its code, while the most interpretable explanation might be a simplified, abstracted, and potentially misleading description of its behaviour. 

However, both reliability and interpretability
are often agnostic to their full extent. For example, \gls{xai} designers often disregard their intended receivers. 
Explainability algorithms need to determine who their receiver is in order to avoid mechanically reporting the same information. 
Lacking knowledge about the receiver makes interpretability a challenging topic. When considering explanations as a causal relationship between some input and output, if the explainee has no understanding of the input (\eg overengineered features), the explanation will become irrelevant~\citep{lipton_mythos_2017}. 

When considering such questions, one should fall back on the most pragmatic one: \textit{What is explainability used for?}
Regardless of context and the nature of the source of explanations, an explanation can be helpful for four potential objectives~\citep{adadi_peeking_2018}: for the sender to \textit{justify} behaviours so that the receiver understands it and to hold accountability, responsibility and transparency; for the receiver to \textit{control} and correct the sender's model via locating flaws and vulnerabilities or to debug; for the sender to \textit{improve} based on feedback from the receiver, such as inspecting nonsensical behaviours and increasing rationality; and for the receiver to \textit{discover} or learn what knowledge the sender has, and how it leverages it to their advantages. 

As such, any desirable \gls{xai} algorithm is tackling at least one of these objectives~\citep{miller_explanation_2019,lipton_mythos_2017,adadi_peeking_2018} while holding some notions (often implicit) of the desirability of explanations related to some of Grice's maxims. 
When performing explanations over models which can be easily accessed, this task is already complex enough.

However, in an era where models are increasingly opaque, auditing models relies on the goodwill of developers to publish their data sources, design principles, and models, as well as to make the tools for auditing available to the community~\citep{chen_how_2023,hassija_interpreting_2024}.
When this is not the case, validating a model as a user becomes unachievable. We, as a community, need better tools to tackle this problem~\citep{longo_explainable_2023}.

This is particularly the case for autonomous agents~\citep{carbonell_is_1997} that interact in an environment: it is tough to understand an agent's purpose or assumed intentions, especially if one has no access to the model or it is opaque.  
This is even harder when the auditor has no access to its reward function (in the case of reinforcement learning (RL) agents) or if the agent is not entirely rational. 
In these cases, obtaining explanations becomes an exercise in anthropomorphism, where a human interpreter attributes behaviours (based on what a human would do, as shown by~\citep{heider_experimental_1944}) in a qualitative analysis that may be inaccurate and risks self-deception and harm~\citep{wortham_what_2016,sartori_sociotechnical_2022}.
Turning such types of analysis into quantitative, verifiable, and reliable explanations will increase the trustworthiness of AI-based systems by having the \emph{explainee} be aware of the quality and manner of explanations provided and have ways to compare them.

\begin{figure}[H]
    \centering
    \includegraphics[width=\textwidth]{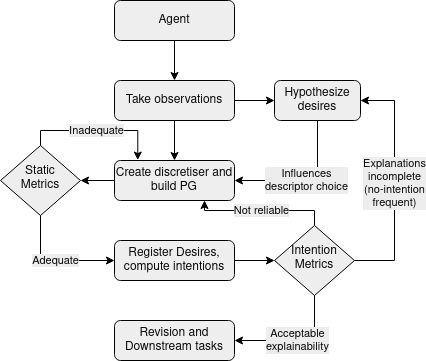}
    \caption{Proposed workflow for extracting explainability. First, (partial) observations of the agent interacting in the environment are taken. The explainee then proposes a (several) discretiser(s) to describe the states, following the heuristics in Section~\ref{sec:design_heur}, that is written in a code they can understand, and that allows them to check a set of hypothesised desires of the agent as described in Section~\ref{sec:desiresintentions}. Then, the resulting \gls{pg} can be evaluated with the metrics proposed in Section~\ref{sec:static_metrics}, allowing the user to gauge the complexity of the \gls{pg} representation and a first estimand of the interpretability and reliability of the model, and can loop back to check a different representation if the equilibrium is not acceptable. \\ Finally, the explainee introduces hypothesised desires into the \gls{pg}, from which they can obtain metrics that validate these hypotheses and give direct estimands of reliability and interpretability, as described in Section~\ref{sec:intention_metrics}. If the explanations are insufficient, the user can filter the regions without apparent intention to hypothesise new desires, as described in Section~\ref{sec:revision}. If the frequency of intentions is too low, the representation may be too complex and can be redesigned. If the results are acceptable, the resulting \gls{pg} can be used for new downstream tasks, such as QA explainability as described in Section~\ref{sec:explanation_algorithms}.}
    \label{fig:graphical_abstract}
\end{figure}

This paper is structured as follows. First, we analyse the state of the art on different types of agent explainability in \S~\ref{sec:background} and we motivate using \gls{pg} as the base method. In \S~\ref{sec:use_case} we briefly introduce the example scenario to be used in the rest of the paper, and then in \S~\ref{sec:methodology} we propose a workflow for creating \textit{post-hoc\/} explainable \gls{pg}-based models of an agent's behaviour by extending previous attempts~\citep{hayes_improving_2017,liu_novel_2023,tormos_llorente_explainable_2023,domenech_i_vila_explaining_2024} 
to achieve better, more interpretable results. 
This method requires no access to the agent program or model, instead it relies on (potentially partial) observations over actions and states reached by the agent, without needing access to reward function, internal state, or design criteria.
The extensions and tools provided are presented in \S~\ref{sec:methodology} and are threefold: enabling a pipeline for verification of human interpretation of agent behaviour via the introduction of teleological explanations of desires and intentions (see Figure~\ref{fig:graphical_abstract} and \S~\ref{sec:design_heur} and~\ref{sec:desiresintentions}); 
using these to provide metrics on interpretability and reliability of the explanations provided (\S~\ref{sec:metrics}); 
and creating algorithms that take into account a shared code that depends on the explainee to answer questions such as ``\textit{Why} did you take a certain action'', ``\textit{How} do you plan to achieve something'', or ``\textit{What} do you plan to do'', which can be composed to get answers at different levels of the causal chain. 
We showcase examples in which these tools can be applied to \textit{justify}, and \textit{discover} agent behaviour, and opportunities to \textit{control} and \textit{improve} it.
In addition, we introduce a hyper-parameter for \textit{commitment}, which allows us to tune the trade-off between the reliability of explanations and the interpretability of agent behaviour overall. 
In \S~\ref{sec:revision}, we explore how a human can use the outputs of the method to improve the quality of the policy graphs produced. In \S~\ref{sec:experiments} we present empirical results of the proposed methodology applied to the example use case, and finally in \S~\ref{Discussion} we discuss our main contributions, possible future work and known limitations of the approach.

Having the capability to produce explanations in these conditions will enable further downstream tasks, such as collaboration and/or competition in multi-agent systems, human collaboration, and especially auditing of such systems~\citep{schaefer_communicating_2017,hayes_improving_2017,tabrez_improving_2019}.

\section{Background}
\label{sec:background}

As mentioned in \S~\ref{sec:introduction}, our focus in this paper is on methodologies for explaining the behaviour of \emph{unknown} agents: agents that are opaque or that have a behavioural policy or model that cannot be inspected. From now on, we will assume that we can only (partially) observe their actions and the environment states. Additionally, we will assume that we have access to a (potentially incomplete) notion of what the desirable behaviour should be in terms of what is needed in order to control, improve or justify the actions of the agent~\citep{longo_explainable_2020,adadi_peeking_2018}, from an explainee point of view. 

\subsection{Agent Explainability}

On the topic of agent explainability, there are a few surveys that enumerate, categorise and analyse the different existing methods and methodologies~\citep{adadi_peeking_2018,puiutta_explainable_2020,arzate_cruz_survey_2020,zhou_evaluating_2021,milani_survey_2022,aha_explainable_2024}. 
One way to categorise explainability methods is to distinguish them based on the time of information extraction, \textit{i.e.} between those that are intrinsic and those that are \textit{post-hoc}~\citep{adadi_peeking_2018,puiutta_explainable_2020}. Intrinsic methods build models that are inherently interpretable or self-explanatory during the design or training of the agent's policy. \textit{Post-hoc} methods, on the other hand, focus on building the explanations by analysing a policy that is already implemented or trained. 

Related to the intrinsic/\textit{post-hoc} categorisation, it is also possible to classify explainability methods into model-specific and model-agnostic~\citep{adadi_peeking_2018,puiutta_explainable_2020}. 
The former are tailored to a specific model or family of model, while the latter methods aim at being able to be used for any kind of agent policy. Most of the approaches found in the literature are model-specific, either by having access to a full or approximate model of the agent or directly designing it~\citep{fox_explainable_2017,albrecht_autonomous_2018,hoshi_why_2018,calvaresi_agent-based_2020,madumal_explainable_2020,winikoff_evaluating_2023,rodrigues_towards_2023,langley_explainable_2024} or by possessing knowledge about specific important parts of the agent's design, such as the reward function~\citep{gyevnar_causal_2023} or the internal task decomposition~\citep{ciatto_towards_2019,verma_discovering_2022}.

Another possible categorisation deals with the scope of each explanation~\citep{adadi_peeking_2018,puiutta_explainable_2020}: whether the method explains the entire behavioural model of the agent and therefore it offers global explanations; or rather it offers local explanations in the sense that they target a specific decision. 
That is, global explanations help explain the model, while local explanations help explain a specific decision~\citep{du_techniques_2019}.
There is another aspect that can be taken into account when characterising an explainability method which is the part of the agent's architecture that should be explained~\citep{milani_survey_2022}. Feature importance methods target quantifying the influence of the features of the agent's inputs (\eg sensory information and percepts) on the decisions made. Learning process methods bind the decisions to specific components of the design or training method that led to the policy, such as the reward function, the Markov decision process or the datasets used. Meanwhile, Policy-level methods try to build a model of the long-term behaviour of the agent.

For the purpose of our work and given the initial premises that define its scope, we propose to focus on methodologies that are:

\begin{itemize}
    \item \textit{Post-hoc}, so that no assumptions need to be made about the design or training process.
    \item Model-agnostic, in order to be able to analyse external opaque agents.
    \item Global and local, as we have two objectives: (1) producing a stable comprehensive model of behaviour~\citep{hayes_improving_2017}, and (2) allowing explanations of particular action decisions tied to long-term processes. 
    \item Policy-level, as we care not only about the reasons for a particular behaviour, but also about the relationship between the behaviour and the environment~\citep{milani_survey_2022}.
\end{itemize}

In order to explain an agent's behaviour, it is necessary to understand which action the agent takes in a state and for what purpose in the context of a trajectory, and not just merely the reasons behind a specific isolated decision.
In most cases, this requires an understanding of the environment in which the agent exists.
In our work, we focus on \textit{policy graphs}, which is a \textit{post-hoc}, model-agnostic, and policy-level explainability method~\citep{hayes_improving_2017,liu_novel_2023,domenech_i_vila_explaining_2024}. Interestingly, this method allows for both global and local explanations. 

\subsection{Policy graphs}

A \gls{pg} (\textit{policy graph}) is a domain model comprising agent and environment behaviour by learning the agent policy (as $P(a|s)$ or probability to choose a certain action $a$ when in a certain state $s$) and the environment's response to agent actions ($P(s'|a,s)$ or probability to end up in a state $s'$ when $a$ is performed in state $s$, often called \textit{world model}~\citep{freeman_learning_2019,gaon_reinforcement_2020,robine_smaller_2023} in the context of sequential decision-making processes). 
However, learning these two distributions is a complex endeavour, as the state space and/or the action space can be large and of varying complexity and/or require state memory to make decisions (\ie it is not enough to know $s_t$, but also $s_{t-1}$ and so on). 
More so, obtaining explainable outputs from a continuous space can be complex, and obtaining reliable estimators of the policy and environment is challenging. 
One common way to simplify this problem is to make the state space finite, discretising real states into more straightforward descriptions.
This simplification allows the \gls{pg} to be a graph-like representation, in which vertices correspond to discrete states and edges correspond to transition probabilities ($P(s_{t+1},a|s_t)$).

A way to solve both problems is to discretise each potentially complex state or action by introducing predicates that summarise states (and potential actions), thus obtaining a discrete, finite number of possible states.
This allows for easy modelling of both probability distributions through frequentist approaches~\citep{hayes_improving_2017}. 
In addition, the usage of human-defined predicates allows for easily interpretable states, which
are then used to provide natural language answers to queries such as identifying conditions for actions (\textit{When do you do $a$?}), explaining differences in expectation (\textit{Why did you do $a$ in state $s$?}), and understand situational behaviour (\textit{What will you do when $X$ is given?}).  
However, the answer to these questions is permanently restricted to immediate results, as it neither provides answers to long-term action behaviour and is agnostic to the agent's goals, desires, or values.

Another way of computing a \gls{pg} can be through automatic discretisation by employing decision-tree approaches to distinguish between continuous states by the difference in actions taken~\citep{liu_novel_2023}. The use of an automatic discretiser simplifies the transformation of the state space into a finite, manageable set. 
However, the predicates that are automatically produced may not be explainable themselves (\eg having `sugar\_high' and `sugar\_low' predicates, distinguished by an arbitrary threshold defined by the decision tree).
Nevertheless, this approach can be used to discover state-regions with consistent agent behaviour (\ie always performing the same action), which are called critical states~\citep{liu_novel_2023}, and also for generating natural language answers to the same questions above.

Similar approaches using predicates have been also used for agents that follow a clear, sequential decision-making process (SDM) towards achieving their goals. Some works~\citep{verma_discovering_2022,verma_autonomous_2023,das_state2explanation_2023,gyevnar_causal_2023} advance on this approach, where agent behaviour is modeled as a series of steps or plans. Unlike SDM-based methods, however, policy graphs do not assume any specific internal model for the agent or its decision-making process. This makes them more adaptable for scenarios where agents might have multiple goals or where their decision-making is not solely goal-oriented. This flexibility is crucial for understanding agents whose behaviour does not necessarily follow a straightforward path or it cannot simply be assumed due to opaqueness.

In previous work~\citep{tormos_llorente_explainable_2023,domenech_i_vila_explaining_2024}, the state of the art on \gls{pg}s is extended in order to cover multi-agent situations in which an agent trained with reinforcement learning cooperates, either along with another \gls{rl} agent, or along with an agent trained to imitate a human player. 
An interesting consequence of the methodology is the creation of surrogate agents~\citep{domenech_i_vila_explaining_2024}: agents that enact policies automatically derived from the generated \gls{pg}. These agents have a comparable behaviour w.r.t. the original trained agent, and therefore this method allows to have policies that mimic the original policy while being transparent. 
This is a form of surrogate agent modelling such as those traditionally used for opaque machine learning models~\citep{adadi_peeking_2018}.

\subsection{Intentionality}
The language explanations provided in the models discussed are limited to locating predicates of the representation relevant to the atomic action selection, which
is not the kind of explainability humans tend to seek~\citep{malle_which_1997,malle_attribution_2022}. 
Instead, the explanations that maximise interpretability over agent behaviour are generally related to understandable end-goals, desires, or rewards, be that explanations regarding why an action contributes toward an objective, why the objective came to exist, or which affordances contributed to achieving an objective. 
It becomes apparent that notions of the agent's objectives and targets are necessary to achieve \textit{good\/} explanations, potentially requiring algorithms inspired by theory-of-mind~\citep{ho_planning_2022,gimenez-abalos_why_2024}.
More so, interpretations incorporating elements of trajectories make agent behaviour more predictable. Trajectories can be defined as sequences of action-state pairs that describe the behaviour of an agent (\eg the trajectory: \emph{I boil water, then I cook the pasta, then I add sauce to produce pasta carbonara} is  predictable because a pattern might have been observed that identifies putting something to cook a very likely action after putting water to boil, etc.).

Given the control, justify, improve framework~\citep{longo_explainable_2020,adadi_peeking_2018}, behaviour predictability is important for producing relevant explanations. One substantial approach to achieve this predictability is by analysing intentionality~\citep{malle_folk_1997,perez-osorio_adopting_2020,dazeley_levels_2021,gimenez-abalos_why_2024}. Intentions are mental states different from other states such as beliefs, desires, knowledge or emotions. The content of an intention is a state of affairs that will be the aim of the agent and to which it commits~\citep{cohen_intention_1990}. However, especially when dealing with opaque agents, intention attribution can be dangerous, so there is a \textit{burden of attribution}. This attribution may not be completely right from a formal perspective~\citep{wright_explanation_2004}, but it is practical and beneficial to do so -- as humans do this attribution process constantly to explain affairs, its burden can be ignored. 
The topic of intentionality and how to deal with intentions and their attribution from a practical point of view will be developed in detail in \S~\ref{sec:desiresintentions}.

\section{Use case}
\label{sec:use_case}

To verify and test the pipeline proposed, several agents of different kinds are analysed in the environment of Overcooked~\citep{carroll_utility_2020}. 
This environment is a \gls{ma} \gls{rl} environment, in which two agents must collaborate to produce and deliver as many dishes as possible in an allotted time. 
The collaborative nature of the environment delivers the possibility of several emergent behaviours beyond what can appear in single-agent environments, and it is particularly interesting from the standpoint of explainability.

The Overcooked-AI environment allows for several layouts and arrangements that motivate the agents' different optimal strategies and behaviours. 
Therefore, we can obtain relevant insights by producing policy graphs for agents trained for each layout and comparing them using the static and intention metrics. 

\begin{figure}[H]
    \centering
    \includegraphics[width=\linewidth]{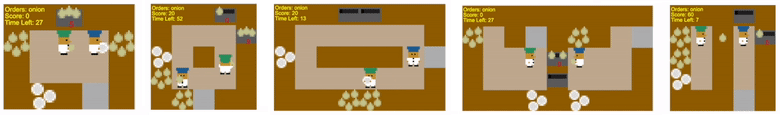}
    \caption{Overcooked visualisation of the analysed layouts, from left to right \simple, \randomone, \randomthree, \unident, and \randomzero}
    \label{fig:layouts}
\end{figure}

This environment is versatile and can target several tasks, layout arrangements, and affordances. The five more used layouts are considered for displaying the \gls{pg} usage and our proposed metrics.
All layouts consist only of the delivery of onion soup. An agent can achieve such an objective by adding three onions to a pot, and after some time steps, the pot will contain soup. 
An agent can collect the soup with a dish and deliver it in a specific `service' tile. Figure~\ref{fig:layouts} is a graphic visualisation of these environments.

Each agent in the environment occupies a tile in a 2D grid-like map,
and faces towards a direction. Two agents cannot occupy the same tile. Agents have six possible actions: 

\begin{itemize}
    \item Moving in one of the four directions (therefore four possible moves) changes the direction they face and, if the tile in that direction is not occupied, moves them to that position. The confrontation is resolved stochastic if two agents attempt to move to the same position.
    \item Interacting with the element in front. This action encompasses several possible actions depending on the context: picking up an item, putting the item in the agent's hands on a counter, putting an onion into a pot, using a dish to pick soup from a cooked pot, or delivering the soup to the service area.
 
    \item Staying, which does nothing and lets the time-step pass.
\end{itemize}

Each of the layouts has its unique strategies that may benefit (or even require) agents' collaboration for achieving result
\begin{itemize}
   \item {\simple} is a cramped room where agent positioning may hinder the other agent. It has a single pot, unlike the rest of the layouts.
    \item {\randomone} and {\randomthree} require agents' coordination to avoid getting stuck in thin corridors. With the longer table in {\randomthree}, agents would benefit from passing onions over the counter.
    \item {\unident} has each agent in different isolated regions, and each side has a different distance between affordances. Agents would benefit from specialising (left agent for servicing, right agent for cooking).
    \item {\randomzero} similarly has each agent in different isolated regions, but each affordance is different, forcing collaboration. The agent on the left needs to pass onions and dishes over the counter to the agent on the right.
\end{itemize}

\section{Methodology}
\label{sec:methodology}

\gls{pg}s are not an out-of-the-box method, as they require some external designing to create and validate the code in which states are described, as well as manual verification of the correctness of the technique. 
We frame the approach towards defining a \gls{pg} in two main designing choices: creating a code for describing states and then formalising hypotheses over the agent's believed desirable behaviour in that code. 

Firstly, a representative sample of observations of the target agent acting in the environment must be collected.  We recommend storing all available information prior to its discretisation, as the pipeline may encourage the designer to change the discretiser: the questions posed by the explainee, or the explainee themselves, can change over time.
In case this is practically impossible (\eg original states or trajectories are too spatially inefficient to store), trajectories should be stored as expressive as possible so that as many different discretisations can be applied \emph{a posteriori}.

Once this information is obtained, a base, non-intentional PG is created by computing and storing probability distributions: $P(s',a|s)$ and $P(s)$; that is, the probability distribution of being in a discretised state $s$, and the transition probabilities when in that state - what the agent does, $a$, and what happens to the state, $s'$. Figure~\ref{fig:graphical_abstract} provides a depiction of our proposed workflow.

\subsection{Policy Graph construction and design heuristics}\label{sec:design_heur}

A \gls{pg}'s construction relies directly on observing an agent's behaviour and discretising it into the discrete state space. 
Any formalism is acceptable for the representation of the internal state representation, but the following properties are greatly encouraged:

\begin{itemize}
    \item The state space is a metric space where we define a distance function
    that computes the similarity between states.
    Generally, this is done with a simple count of different predicates~\citep{hayes_improving_2017}, but more sophisticated approaches that account for predicate semantics could provide \textit{better\/} explanations.
    \item The resulting state space is small enough that the agent can map states from new observations to existing, already observed states.
    \item The resulting state representation is interpretable to a human or downstream task, who can understand the original state's properties based on its discretised version's internal representation. This understanding can be incomplete; it is only enough to justify or interpret agent behaviour based on it.
    \item The resulting state representation allows to formally represent \textit{desires} as introduced in \S~\ref{sec:desiresintentions}. This step requires parallelising the process of designing the \gls{pg} and hypothesising over desired behaviour, as the ability to test a desire depends on representing it for discrete states.
\end{itemize}

The rationale for these heuristics can be understood from the trade-off between interpretability and reliability.
On the one hand, the first two properties are for increasing reliability. The probability distribution only represents the real world if observations are few to appear frequently in the graph. 
In addition, by introducing a notion of distance, one can consider the state-space a metric space and use similarities between states to compensate for the lack of observations at the cost of some reliability.
On the other hand, the representation of the internal states will be part of the code shared between the explainee and the model. 
If such code is not shared, the result will hardly be interpretable. This, in turn, allows for explanations that conform to what the explainee can understand.

When merging both necessities, it is noticeable that they go in opposite directions: having a small state-space hinders having the expressivity demanded by an extensive code of communication between the explainee and the model, thus hindering interpretability. 
Similarly, a thorough state description implies a more extensive state-space, in which the specificity of each state will result in a lower probability of reaching it in our observations, lowering the reliability of the probabilities conditioned to being in such a state.
This is a significant problem when working with real problems with scarce data available, as it requires finding a state representation slim enough that all states are sufficiently observed to produce explanations. 
This complexity also explodes when considering that the critical states requiring explanations are often less frequent, thus increasing data-gathering requirements.

Handling the trade-off between interpretability and reliability depends on the task at hand, thus requiring metrics for evaluating which of the two sides is favoured by a specific discretiser or representation.

Finally, although any person, including non-experts, can propose discretisers, their usefulness relies partly on the state-space description. 
Experts on the field are more likely to correctly guess which environment parameters are more relevant to the agent's behaviour and thus be more efficient in their search for the optimal discretiser, but the metrics proposed in \S~\ref{sec:metrics} and the pipeline described in Figure~\ref{fig:graphical_abstract} allow non-experts to bridge the gap through more iterations of the process.

Following previous work~\citep{cortes_testing_2022,tormos_llorente_explainable_2023}, we pick a simple discretiser and distance that are directly matched with our representation. 
We describe each state using problem-specific propositional logic predicates, discretising real states by evaluating the truth-value of each predicate and assigning the equivalent discretised state. 

We take the number of different predicates between two representations with no weighting for distance. We note that more sophisticated representations exist, such as employing decision trees~\citep{liu_novel_2023},
using clustering on state CLIP embeddings, or even Scene Graphs. For the problems tackled in this article, the most straightforward approach worked well enough.

\subsection{Explainability based on desires and intentions}\label{sec:desiresintentions}

Most explainability algorithms in the literature focus on establishing some causal relationship, correlation, or \textit{relevance\/} between some input variable and the model's output~\citep{lundberg_unified_2017,ribeiro_why_2016,selvaraju_grad-cam_2017}.
However, when asking a human why they put a cooking pot on the hob, it is arguably the case they will reply:
\textit{Because the pot was full of water and the hob was not being used}. 
A correlation may exist between a pot full of water and the cook placing it on top of the hob, as cooks often fill the pot with water when they plan to boil it. 
However, the motivator of such behaviour is not the availability of the pot and the hob but the intention of the task.
As humans are capable of \emph{consciously} setting themselves goals to pursue, explanations involving human intent are often teleological, including or relating to the ends of the behaviour (\eg \textit{because I want to cook some pasta}). In many cases, these teleological explanations encompass the realms of morals, ethics and politics~\citep{wright_explanation_2004,johnson_teleology_2005}, but the actual intention acts as the main predictor of the existence of abstract mental states such as holding a particular value or moral norm~\citep{godin_bridging_2005} (\eg self-preservation).

In our example, an explanation a human cook would give to someone who does not know how to cook would more likely be: \textit{Because I am making pasta carbonara, and for that, I need to cook the pasta, and for that I need to boil water}. Although further explanations may involve state variables such as the state of the pot or the hob, the natural communicative act cannot constrain itself to that level alone~\citep{winikoff_evaluating_2023}.

When analysing a (reasonably well-performing) agent's behaviour in a domain, humans tend to anthropomorphism~\citep{heider_experimental_1944,wortham_what_2016,sartori_sociotechnical_2022}.
So long as the agent's actions are not entirely random and there is a way to establish logical inferences from them from a teleological perspective~\citep{searle_intentionality_1980,wright_explanation_2004}, humans 
attribute intentionality to the agent (\eg \textit{It has grabbed the onion because it intends to put it in the pot later on\/}).
This is especially the case for most toy environments (\eg games) of which the human observer has some knowledge of how to solve and thus is expecting certain behaviours of its virtual homologous, and it extends to experts observing agents' behaviour in their domains~\citep{somers_how_2018,park_how_2022}.

However, when observing a low number of interactions, such attributions are subject to anecdotal fallacy unless systematically verified over many interactions. 
In this section, we present a way to leverage this cognitive bias in order to enable agent explainability to answer the \textit{what}, \textit{why}, and \textit{how} questions in a manner not dissimilar to how a human would. 
This is done through the introduction of \textit{agent desires}, which can be modelled in diverse ways, and \textit{agent intentions}, that is, the desires we expect the agent to accomplish (soon) as allowed by the environment~\citep{cohen_intention_1990}. In addition, we introduce to this pipeline a hyper-parameter that directly lets the human control the interpretability-reliability trade-off: the commitment threshold.

\subsubsection{Desires}\label{sec:desires}
In this work, \textit{desires} are introduced as hypotheses over expected behaviour: the work of anthropomorphism by a human observer that has some rudimentary or expert knowledge of the task the agent is solving. 
This desire may or may not express itself in the behaviour of the agent, and thus they require verification. 
If a desire truly expresses itself, it is often due to the design concerns through which the agent was created, be that some particular rule in the system, the design of a reward function, or a statistical bias in the data it trained on.

Pragmatically, defining a desire requires understanding when it is fulfilled. We distinguish between several cases such as reaching or staying (achievement and maintenance goals respectively, as shown by~\citep{van_riemsdijk_goals_2008}) in states where some qualities hold (\eg in Cartpole, to stay in a state where the rod is upright), to execute an action in such states (\eg in Overcooked, to interact with the service zone with soup on my hand), or performing a particular transition between world states (\eg in racing, crossing the finish line). 
These also extend to their negative forms, such as `not' staying in some states.

We concentrate on the second type: action-focused. Many desires can be reduced to this kind with clever discretisation~\citep{domenech_i_vila_explaining_2024}, but extending the framework to those that cannot is easy. 
Action desires can thus be defined as a tuple $\langle S_d, a_d \rangle$ containing a discrete state region ($S_d = \{s\in S| s \vdash d\}$, where $s \vdash d$ means that the state satisfies the desire's condition), and the action $a_d$ that would be desirable in such state region. 
As the explainees themselves provide this characterisation, they are expected to understand it when it becomes the \textit{finality\/} of explaining behaviour.

\begin{figure}[htp]
    \centering
    \begin{subfigure}[b]{0.49\textwidth}
            \centering
            \includegraphics[width=\textwidth]{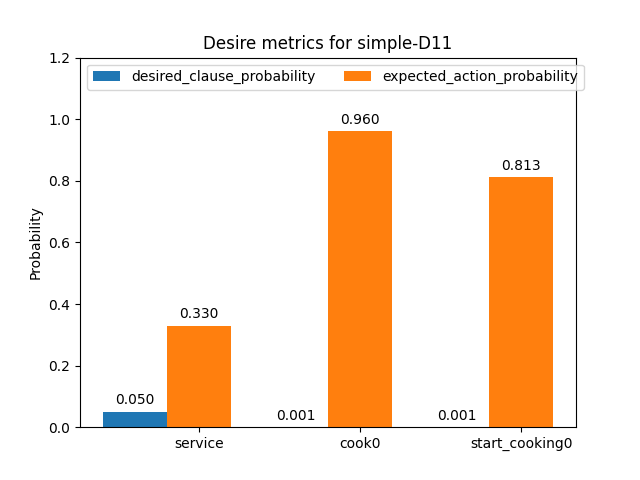}
            \caption[]%
            {{\small {\cmasone} in Environment {\simple}}}    
            \label{fig:desire_cmas_simple_11}
        \end{subfigure}
        \hfill
        \begin{subfigure}[b]{0.49\textwidth}  
            \centering 
            \includegraphics[width=\textwidth]{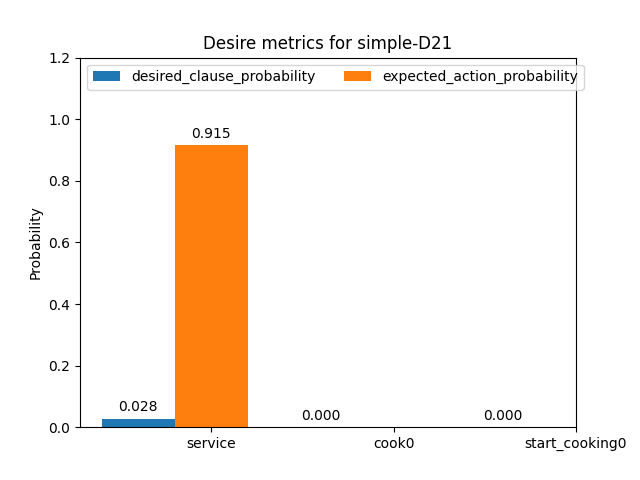}
            \caption[]%
            {{\small {\marcone} in Environment {\simple}}}    
            \label{fig:desire_marc_simple_11}
        \end{subfigure}
        \begin{subfigure}[b]{0.49\textwidth}
            \centering 
            \includegraphics[width=\textwidth]{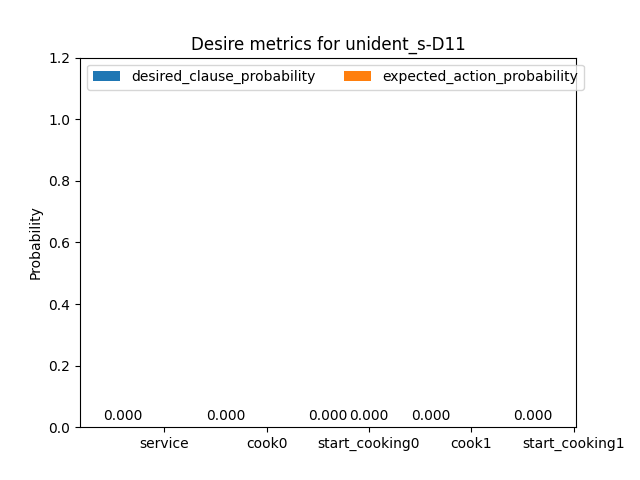}
            \caption[]%
            {{\small {\cmasone} in Environment  {\unident}}}    
            \label{fig:desire_cmas_unidents_11}
        \end{subfigure}
        \hfill
        \begin{subfigure}[b]{0.49\textwidth}   
            \centering 
            \includegraphics[width=\textwidth]{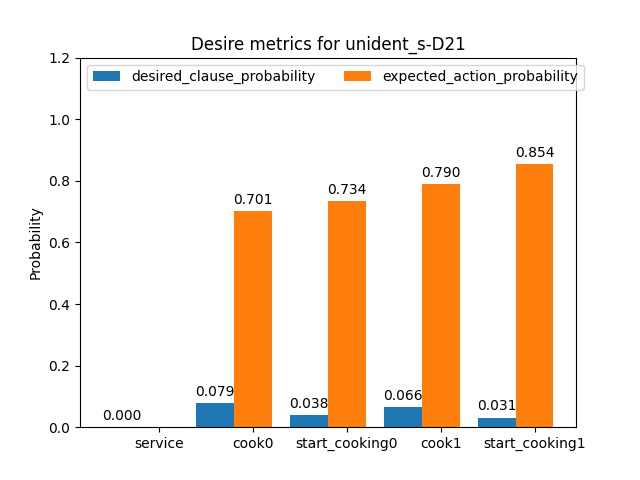}
            \caption[]%
            {{\small {\marcone} in Environment  {\unident}}}    
            \label{fig:desire_marc_unidents_11}
        \end{subfigure}
    \caption{Desire metrics for two types of agents ({\cmasone} and {\marcone}) in two environments (\simple~and {\unident}) environments and the same discretiser (1), all described in Section~\ref{sec:experiments}. The desire probability (blue) is very low for all cases. Higher values of desire probability are indicative of higher performance, subjected to the desire being actually fulfilled (orange). {\marcone} {\unident} never fulfills the service desire, but is quite frequently fulfilling the rest. Note how {\cmasone} is never in a state in which it can fulfill any hypothesised desire in {\unident}, meaning its behaviour is unexplainable.}
    \label{fig:desire_metrics}
\end{figure}

Calculating relevant information over these desires is trivial under the probabilistic description of a \gls{pg}. \textit{How likely are you to find yourself in a state where you can fulfil your desire by performing the action?} can be computed as the desire state region probability $P(s\in S_d) = \sum_{s\in S_d} P(s)$). \textit{How likely are you to perform your desirable action when you are in the state region?} can also be computed as $P(a_d|s\in S_d)=\sum_{s\in S_d} P(a_d|s)*P(s)/P(s\in S_d)$). 
These metrics can be found for some of the experimental environments in Figure~\ref{fig:desire_metrics} 
\footnote{The description of each desire can be found at the end in this section.}, and they serve as a first verification of the desires. Each graph represents an agent's desires, evaluating the same desire for each agent. Except for the first one (\cmasone), at least one of their desires is shown not to exist, as the desirable action is never performed in the state region, illustrated by the lack of expected action probabilities.

This addition is not a panacea for the problem. Most states in a problem do not manifest the specific conditions for immediately fulfilling a desire, as $P(s\in S_d)$ is expected to be low in most cases. 
The reliability of the obtained metrics is directly measurable by $P(a_d|s\in S_d)$ (\ie explanations expressing that the cause of a certain behaviour is that the agent willing to fulfil the desire can be wrong if the action of the desire is not performed).

That being the case, given that only states in a desirable region can be interpreted --and those states often account for a very small slice of time-- the agent's behaviour cannot be safely interpreted most of the time. 
For this purpose, intentions are introduced in \S~\ref{sec:intentions} as an extension of this framework.

Using the case of Overcooked, which is further described in \S~\ref{sec:use_case}, as an example, the following desires are \textit{guessed} and tested, formalised using propositional:

\begin{enumerate}
    \item The agent desires to service soup: The state region is all states where the agent can deliver soup (that is,
    all states where the agent has soup and the service zone is in the interact position), and the action to be performed is to interact.
    \item The agent desires to cook: The state region is all states where the agent can add an onion to a pot with already one onion in it (\ie having an onion, the pot being in the \textit{preparing} state, and the pot being in the interact position), and the action to be performed is to interact.
    \item The agent desires to start cooking: Analogous to the desire to cook, but the state region requires the pot to be \textit{empty} instead of \textit{preparing}.
\end{enumerate}

When proposing these desires at a first iteration, the intention was to seek high-granularity tasks in order to verify the explainability of the system on a small subset of desires. 
More desires could be formulated, such as the desire to grab an onion when the pot is empty or preparing, but these were enough to achieve good interpretability metrics. 

\subsubsection{Intentions}\label{sec:intentions}
In order to extend explanations to the keyword of \textit{why}, the transitional information of a \gls{pg} can be leveraged. An agent's intention to fulfil a desire exists if it can be fulfilled (given by world dynamics and its understanding), and the agent commits to doing so~\citep{cohen_intention_1990}. 
Our empirical observations of the agent's behaviour capture both requirements. 

Loosely defined, intentions of fulfilling a desire $I_d(s)$ can be measured by considering the probability that the agent will attain the desire from a given state. 
Informally, it is the sum of probabilities of all possible paths starting in one state that arrive at any state where the agent can fulfil the desire and is fulfilled.

Formally, let $\mathcal{P}(s,d)$ be the (potentially infinite) set of paths starting from $s$ and arriving at any $s' \in S_d$ (not counting paths that fulfil the desire midway through). 
The intention of such a desire can be thus computed as:
$I_d(s) = \sum_{p\in \mathcal{P}(s,d)} P(a_d|last\_state(p))*P(p)$, where $P(p)$ is the probability of traversing path $p$ as computed by the \gls{pg}: $P(p)=\Pi_{s',a,s_t\in p} P(s',a|s)$. 
One could consider the metrics used to describe desires to be myopic intentions restricted to paths of 1-action length.

Given the potentially infinitely-looping paths, the computation is done backwards, starting from $S_d$ and recursively propagating intention updates to the parent states. 
A stopping criterion $\epsilon$ is introduced to stop the propagation of intentions below a certain probability. 
A complete description of the algorithm can be found in Algorithms~\ref{alg:desire_registry} and \ref{alg:int_propagate}.

\begin{algorithm}
\caption{Register a Desire into a PG and propagate intentions}\label{alg:desire_registry}
\begin{algorithmic}
\Require $d, PG$
\For{$s \in PG$}
\State $I_d(s) \gets 0$
\EndFor
\For{$s \in S_d$}
\State $increment \gets P(a_d|s)$
\State $\texttt{Propagate\_intention}(s, d, PG, increment)$
\EndFor
\end{algorithmic}
\end{algorithm}

\begin{algorithm}
\caption{Propagate intentions to node $s$. Propagation of desires is stopped from crossing through the transitions that would fulfil them, as not doing so would compute the `expected number of times a desire will be fulfilled' instead (which can be above 1).}\label{alg:int_propagate}
\begin{algorithmic}
\Procedure{Propagate\_intention}{$s, d, PG, increment$}
\State $I_d(s) \gets I_d(s)+increment$
\For{$p \in \{p\in PG| P(S'=s|S=p)\neq 0\}$}\Comment{All parents of s}
    \If{$p \notin S_d$} \Comment{P cannot fulfil the desire, all transitions are valid}
        \State $propagable\_intention \gets P(S'=s|S=p)*increment$
    \Else \Comment{P could fulfill the desire by doing $a_d$, don't count those}
        \State $propagable\_intention \gets P(S'=s, A\neq a_d|S=p)*increment$
    \EndIf
    
    \If{$propagable\_intention\geq \epsilon$}\Comment{Stop criterion, usually 1e-4}
        \State $\texttt{Propagate\_intention}(parent, d, PG, propagable\_intention)$
    \EndIf
\EndFor
\EndProcedure

\end{algorithmic}
\end{algorithm}

Introducing $I_d(s)$ as a tool allows the user to ask for complex queries. For example, one could ask \textit{What do you intend to do in state $s$}, to which the agent could reply with all desires with an $I_d(s)$ over a certain threshold. 
Another question could be \textit{Why did you take action $a$ at state $s$?}, to which the algorithm would reply: \textit{I have the desire $d$, which I can bring about from state $s$, and by performing action $a$ either I am closer to achieving it, or there is a chance I will increase my odds of doing so}. 
The algorithms for replying to these queries can be found in \S~\ref{sec:explanation_algorithms}.

The intention value is directly interpretable, as it is the probability that some desire will be brought about given a state. 
However, the lower the intention, the more uncertain its fulfilment, and the continuous property of intentions makes it so that a user may convince themselves of wrong information by vastly overestimating a probability. 
For this, we propose to restrict intentions to being above a parameter called the \textit{commitment threshold} $0<C\leq 1$, which specifies at which minimum probability the explainee is willing to believe the agent will try to fulfil a desire. 
Any $I_d(s)<C$ is to be disregarded, whereas, for any state $s$ such that $I_d(s)\geq C$, the agent can be said to have (at least some) intention to fulfil $d$. we can say that $s$ is attributed to the intention $I_d$.

This \textit{commitment threshold} is a parameter directly related to the reliability-interpretability trade-off. 
When the parameter $C$ takes on higher values, it boosts the likelihood that any state to which intention is attributed will fulfil the desire. 
On the other hand, when $C$ is lower, more states are attributed with intentions, which makes a more significant part of the behaviour interpretable. 
However, some intentions may go unfulfilled, leading to less reliable explanations.

We measure and control this trade-off by extending the desire metrics into `intention' metrics (dependent on $C$), which are introduced in \S~\ref{sec:intention_metrics}: the \textit{attributed intention probability} and the \textit{expected intention probability}. These two metrics which are estimands of interpretability and reliability, respectively, and can be computed for each desire and the \gls{pg} overall.

\subsubsection{Explanation-extraction and answerable queries}\label{sec:explanation_algorithms}

To leverage the computed intentions, one needs to ponder which questions require answering for explainability to make sense and be helpful. To do this, we focus on studies on how human explainers achieve this. 
In the folk-conceptual theory of behaviour explanation, one can categorise between explanations provided for unintentional and intentional behavior~\citep{malle_which_1997,malle_attribution_2022}. 
Most previous work~\citep{hayes_improving_2017,liu_novel_2023,domenech_i_vila_explaining_2024} focuses on answering \textit{why} queries by listing beliefs of the agent, which would fall in the kind of explanations usually provided for unintentional behavior. For example, \textit{I move north when I am south of a delivery area and have the part}~\citep{hayes_improving_2017} is a case of \textit{why} question where \textit{why} means \textit{what caused}.
Other types of questions (and, therefore, answers) must be provided for intentional kind, such as when \textit{why} means \textit{why for}. An example of these would be the aforementioned example of \textit{I boil water because I want to make spaghetti}. 
The design focus on which questions need answers is motivated by two principles: information given should be minimal (following the maxim of quantity), but enough question types and asking methods should be available to extract further information if the current is insufficient (to ensure interpretability).

In previous work~\citep{malle_attribution_2022}, intentional behaviour explanations were categorised into three modes:

\begin{itemize}
    \item Reason explanations, which concern themselves with the causality of an action being taken as assigned to `what the intention is, and how an action favours it', and are by far the most common kind (3 in 4 cases)~\citep{malle_attribution_2022,malle_how_2004,malle_attributions_2007}. 
    In addition, this type of explanation tends to include additional reasons, such as avoiding alternative outcomes or beliefs about the context.
    \item \gls{chr} explanations, which concern themselves with explaining the precursor factors to the reasons it is chosen (including intentions). 
    In reinforcement learning, this is intrinsically, but not exclusively tied to the chosen reward function (\eg emergent behaviour). 
    As an alternate example, in agents with a Belief-Desire-Revision (BDI) architecture~\citep{rao_modeling_1991} the action is tied to the designer's choice of desires, values, or arises as emergent behaviour.
    
    \item \gls{ef} explanations, which concern themselves with explaining why an action which is apparently desirable was successful.
\end{itemize}

To answer all of these queries succinctly, we propose a rewording and partition of these questions, which, when combined, allow the explainee to put together satisfactory answers to all of the above, focusing on Reason explanations: \textit{What} do you intend to do now? \textit{How} do you plan to fulfil it? \textit{Why} are you taking this action now? \textit{Why} are you not taking this other action? 
And finally, \textit{When} do you manifest an intention? We propose algorithms for answering the first three, which suffice for good answers to reason explanations. 
Answering the \textit{when} question would allow us to provide explanations resembling  `enabling factor explanations', whereas \gls{chr} is out of the scope of this paper.

The first question is the easiest one to solve: given a state $s$, returning any attributed intentions $I_d(s)\geq C$.
However, this needs a more satisfactory explanation of the reason. For an intention to exist, the agent needs to have the desire and believe it can be fulfilled. Suggesting the former may not elucidate the latter, and as is apparent by the frequency of reason explanations, it is a prevalent necessity. As an example, consider the Cartpole environment\footnote{\url{https://gymnasium.farama.org/environments/classic\_control/cart\_pole/}}: If an agent returns that it intends to straighten the pole up in a state where it is falling left, we would expect an answer such as the following:
``\textit{My goal is to keep the pole upright. Currently, the pole is upright but leaning to the left, and I am not on the left edge, so I move to the left. This results in a situation where the pole is no longer leaning left, thus achieving my goal\/}''. To get this answer, the second question inquires about the method to achieve it.

To answer the question of \textit{how} they believe the goal will be achieved, and also \textit{why} they believe it can be achieved, we leverage the \gls{pg} knowledge. Algorithms~\ref{alg:how} and~\ref{alg:how_stochastic} return increasingly in-depth answers to the query. 
Intuitively, the former returns the most optimal path to fulfilling an intention by picking the successor (where a successor holds $\{s'\in PG| P(S'=s', a=a|S=s)\neq 0\}$) to the current considered state that holds the most significant increment in $I_d$ until $d$ is fulfilled. 
As the intention in a state $s_i$ is the average of the intentions in $s_{i+1}$, it is always the case that either at least one successor has a larger or equal intention or the current state can directly fulfil the desire. 
This algorithm gives a plausible path but needs to account for setbacks or possible alternatives and is thus only partial.

Algorithm~\ref{alg:how_stochastic} compliments this by considering instead randomly sampling state successors from $P(s',a|s)$, recording multiple paths and classifying them between success and failure, where the former is an arrival at a state such that the action can be fulfilled and the latter is an arrival at some state where the intention is no longer attributed (\ie falls below the commitment threshold).

\begin{algorithm}
\caption{How do you plan to fulfill $d$ from $s$?}\label{alg:how}
\begin{algorithmic}
\Procedure{how}{$d, s, PG$}
\State $current \gets s$
\If{$s \vdash d$} \Comment{State can fulfill desire}
    \State \textbf{return} $a_d$ \Comment{return action that fulfills the desire}
\EndIf
\State $s' \gets argmax_{s', a\in Succ(s)} I_d(s')$ \Comment{Maximum intention possible future state and action}
\State \textbf{return} $\texttt{cat}(a, s', \texttt{how}(d, s', PG)$
\EndProcedure

\end{algorithmic}
\end{algorithm}

\begin{algorithm}
\caption{Stochastic how do you plan to fulfill $d$ from $s$?}\label{alg:how_stochastic}
\begin{algorithmic}
\Procedure{how\_stochastic}{$d, s, C, PG$}
\State $current \gets s$
\If{$s \vdash d$} \Comment{State can fulfill desire}
\State \textbf{return} $a_d, Success$ \Comment{return action that fulfills the desire}
\EndIf
\If{$I_d(s') < C$} \Comment{Intention is no longer attributed in this state, it is below commitment threshold}
\State \textbf{return} $Failure$
\EndIf
\State $s',a \sim P(s',a|s)$
\State \textbf{return} $\texttt{cat}(a, s', \texttt{how\_stochastic}(d,s',C,PG) $
\EndProcedure

\end{algorithmic}
\end{algorithm}

Although these two questions are enough to explain the reasons for having intentions, answering for agent `behaviour' is intrinsically tied to the choice of actions taken and, therefore, must also account for the action perspective. To do this, it is necessary to answer the question of \textit{why} an action is taken.
A way to answer is by considering the possible effects an action $a$ will have in a particular state $s$, grounded in increases of intention that motivate the change. Actions can be broken down into unintentional and intentional. 
This paper defines the latter as `actions that help support further one intention', which in turn mean higher odds of it succeeding, hence an increase in $I_d(s)$ for some $d$.
However, this would not account for risky actions. 
For example, a plausible explanation for participating in a lottery would be gaining money, but the probability of such happening is low. 
An action that can further an intention may also hinder it depending on the following state it achieves (\eg winning or losing). Instead, the interpretation of this answer may need to rely on probabilities of increase and expected increases when taking an action. 

If no attributed desire exists in the state, then the action is apparently unintentional from the point of view of the \gls{pg} and considered desires. 
Else, each attributed desire is a candidate. For each, we compute the expected intention increase when executing the action ($\mathbb{E}_{P(s'|a,s)} I_d(s') -I_d(s) = \sum_{s'} P(s'|a,s)*I_d(s') - I_d(s)$ ). 
If the intention increase is positive, the action is expected to further the intention, which is a good enough explanation (over that desire).  
Else, it can be considered a gamble: computing the probability of positive increase ($P(I_d(s') \geq I_d(s)|s,a)$), and the expected positive intention increase ($\mathbb{E}_{P(s'|a,s,I_d(s')\geq I_d(s))} I_d(s')$). 
The explainee can consider these metrics to gauge how likely the action was to further the intention and by how much. 
A probability distribution function can also be considered, computing $P(I_d(s') - I_d(s)|s,a)$ for visual analysis. If neither metric is acceptable for any desire, the behaviour is also considered unintentional from the point of view of the \gls{pg} and considered desires. 
This behaviour frequently happens when analysing RL agents, as there may exist vestigial exploration behaviour (\,  i.e. trying a priori non-optimal actions to test if there are unexplored possibilities that are better than the current optima).

\begin{table}[htb]
    \centering
    \begin{tabular}{|c|c|}
    \hline
        What? & desire\_to\_service (0.82) \\\hline
        Why (Interact)? & \begin{tabular}[c]{@{}l@{}}
        I want to do Interact for the purpose of furthering \\ desire\_to\_service as it has a 0.99
        probability of an \\ expected increase of 0.01
        \end{tabular}
        \\\hline
    \end{tabular}
    \caption{Answers to What and Why questions in State 84 of agent {\cmasone} in Environment {\simple} using PG-discretiser 1}
    \label{tab:what_why}
\end{table}

\begin{table}[htb]
\footnotesize
\begin{tabular}{|l|l|l|l|l|l|}
\hline
                              Interact(0.82)                                                                                                    & Right(0.89)                                                                                                                                                                              & Down(1.0)                                                                                                                                                                                    & Interact\\ &&& (fulfilled)    \\\hline
                       {\color[HTML]{32CB00} \begin{tabular}[c]{@{}l@{}}HELD\_PLAYER\\\quad (SOUP)\\ POT\_STATE\\\quad (POT$_0$;¬STARTED)\end{tabular}} & {\color[HTML]{32CB00} \begin{tabular}[c]{@{}l@{}}ACTION2NEAREST\\\quad (ONION;INTERACT)\\ ACTION2NEAREST\\\quad (POT$_0$;LEFT)\\ ACTION2NEAREST\\\quad (SERVICE;BOTTOM)\\ ACTION2NEAREST\\\quad (SOUP;BOTTOM)\end{tabular}} & {\color[HTML]{32CB00} \begin{tabular}[c]{@{}l@{}}ACTION2NEAREST\\\quad (ONION;TOP)\\ ACTION2NEAREST\\\quad (SERVICE;INTERACT)\\ ACTION2NEAREST\\\quad (SOUP;RIGHT)\\ POT\_STATE\\\quad (POT$_0$;PREPARING)
                      \end{tabular}}        & {\color[HTML]{32CB00} } \\\hline
{\color[HTML]{CB0000} \begin{tabular}[c]{@{}l@{}}HELD\_PLAYER\\\quad (DISH)\\ POT\_STATE\\\quad (POT$_0$;FINISHED)\end{tabular}}     & {\color[HTML]{CB0000} \begin{tabular}[c]{@{}l@{}}ACTION2NEAREST\\\quad (ONION;RIGHT)\\ ACTION2NEAREST\\\quad (POT$_0$;INTERACT)\\ ACTION2NEAREST\\\quad (SERVICE;RIGHT)\\ ACTION2NEAREST\\\quad (SOUP;RIGHT)\end{tabular}}  & {\color[HTML]{CB0000} \begin{tabular}[c]{@{}l@{}}ACTION2NEAREST\\\quad (ONION;INTERACT)\\ ACTION2NEAREST\\\quad (SERVICE;BOTTOM)\\ ACTION2NEAREST\\\quad (SOUP;BOTTOM)\\ POT\_STATE\\\quad (POT$_0$;¬STARTED)\end{tabular}} & {\color[HTML]{CB0000} }\\\hline
\end{tabular}
    \caption{Answer (deterministically) to the question \emph{How to deliver soup?} from State 84 of agent {\cmasone} in Environment {\simple} using PG-discretiser 1. At each stage, it responds with \emph{what action it would do in the state} and \emph{how it believes the state could change} (both added and removed predicates after applying the action). In green: added predicates; in red: removed predicates. The header row represents: (Action \& $I_d(s')$).}
    \label{tab:how}
\end{table}

Answering all possible inquiries of an explainee would require considering counterfactual explanations. 
These are currently outside 
the grasp of \gls{pg}s, as they require more than statistical knowledge~\citep{pearl_causality_2000}. 
For example, when questioning an agent behaviour, a user with preconceptions over optimal behaviour would ask: \textit{Why did you not take action $a'$ at state $s$ (which I believe to be optimal)?}. 
The closest way to answer this question would be to ask, \textit{Why would you take action $a'$ at state $s$?} as if the action was indeed taken, and ask the same for the chosen initial action to contrast and compare answers. 
However, this runs into limitations, especially if action $a'$ was infrequently (or never) taken at state $s$ during the creation of the \gls{pg}. 
This kind of explanation could be leveraged to improve agent behaviour: if the question is asked and indeed $a'$ was undersampled, an agent may be coerced to test it more often, updating its behaviour and \gls{pg}. Nevertheless, it seems necessary to hold a causal model of actions and world predicates beyond observations to answer these counterfactuals~\citep{pearl_causality_2000}.

Finally, to find enabling factor explanations, a potential avenue would be to answer queries such as \textit{When is an intention for $d$ manifested?} or \textit{What properties does a state $s$ need to hold so that the agent commits to desire $d$?}. 
As of now, intentions are computed by looking at future states, but since the current state properties are what define the probability of arriving at future states, it should be the case that there is a causal relationship between state properties and manifested intentions. 
For example, an agent may manifest the intention to deliver when the pot is finished, regardless of future states.

\subsection{Metrics}\label{sec:metrics}

Several heuristics for \gls{pg} design have been introduced in the previous sections. However, managing these heuristics and achieving the desired balance between reliability and interpretability cannot be a blind task. Much like the intended explanations, the design processes should be quantitatively analysed to make the algorithm available to the user.

For this purpose, in this section, we propose metrics defined as functions that allow us to assess and quantify the performance and effectiveness of our proposed pipeline in terms of the explanations produced. These functions should be effectively regarded as distance functions that enable quantitative comparisons between explanations in a metric space.

Given that the proposed pipeline works in two stages (first constructing a \gls{pg}, and then proposing desires and intentions), the metrics in this section are split depending on which specific part of the pipeline it makes sense to apply them.

Static metrics can be seen in the literature~\citep{cortes_testing_2022,liu_novel_2023}, which take the PG as a probabilistic graphical model and analyse its properties statically. 
Although these are the most used and intuitive, some inherent weaknesses arise. 
One of the sources for this is that no information on the criticality of decision in a state is present on a \gls{pg}, meaning that surrogates have to be taken. We present the limitations of static analysis in a toy experiment in \S~\ref{sec:semaphor}.

However, such a problem can be solved by introducing desires and metrics that leverage their information to compute the reliability and interpretability of explanations. As these metrics require a set of user-defined desires (which can be created iteratively), the guides they provide at early stages may be biased to a sub-optimal representation. 
As such, we propose relying on static and intention metrics, leaning more on the latter as the \gls{pg} is refined.

\subsubsection{Static Metrics}\label{sec:static_metrics}

Static metrics 
analyse the graph's properties regardless of intentions and desires. These allow for an idea of the variability of the expected agent behaviour in different scenarios, which can be helpful to pick the best state representation for the \gls{pg} and compare several ones. 
We consider three approaches to the task, each evaluating different but relevant points: entropy, behavioural similitude, and trajectory likelihood.

Entropy is one of the most natural ways of evaluating how informative the \gls{pg} model is: if knowing the current state unequivocally determines the following action and state, then the \gls{pg} is perfect, the explanations are entirely reliable, and a policy derived from it could substitute the original agent. 
This will only be the case for toy cases, but entropy will quantify how close we are to such an ideal state. 

For the purpose of \gls{pg}s, state entropy is computed as follows:
\begin{equation}\label{Eq:s_g_entropy}
    \entropy(s) = -\sum_{s',a \in \{s',a: P(s',a|s) \neq 0\}} P(s',a|s) * log_2 P(s',a|s)
\end{equation}
This metric can be understood as the expected number of bits necessary to encode the immediate future of the node: the lower, the less uncertainty exists over the agent and environment's behaviour. 
The future of the node may be further decomposed in two factors: action entropy $\entropy_a(s)$ (Eq.~\ref{Eq:s_agent_entropy}), and future state (or world) entropy $\entropy_w(s)$ (Eq.~\ref{Eq:s_world_entropy}), holding that $\entropy(s)=\entropy_a(s)+\entropy_w(s)$.

\begin{equation}\label{Eq:s_agent_entropy}
    \entropy_a(s) = -\sum_{a \in \{a: P(a|s) \neq 0\}} P(a|s) * log_2 P(a|s)
\end{equation}
\begin{equation}\label{Eq:s_world_entropy}
    H_w(s) = -\sum_{a \in \{a: P(a|s) \neq 0\}}  P(a|s) * \sum_{s' \in \{s': P(s'|s,a) \neq 0\}}  P(s'|s,a) * log_2 P(s'|s,a)
\end{equation}

The decomposition of entropy in two parts shows a key insight on the balance for creating a \gls{pg}: a low number of different discretised states results in fewer possibilities for $P(s'|s,a)$ and likely a lower $\entropy_w(s)$, but at the same time it is likely that a state s determines the following action perfectly by $P(a|s)$, and thus lowers $\entropy_a(s)$. 
This equilibrium is also present in the reliability and interpretability side: the more states there are, the more difficult it is to understand agent behaviour, as one has to shift to local state regions to analyse graphs that are too large. 
However, too simple graphs with few nodes show much larger action uncertainty, making the outputs less reliable. It should also be taken into account that the larger the \gls{pg}, the more agent observations should be taken to lower the variance of estimations of $P(s'|s,a)$, or the resulting graph will not be reliable even despite entropy computations.

These entropy metrics can be extended to the full graph by taking the expectancy ($\mathbb{E}(H_x(s)) = \sum_s P(s)*H_x(s)$, for $\entropy(s),\entropy_a(s),\entropy_w(s)$).

In the literature, the mean of entropies has also been observed~\citep{liu_novel_2023} by not accounting for $P(s)$. This can be a desirable change, especially given that, for some problems, taking specific actions may only be critical in certain unlikely states. We show the limitations of entropy as an evaluation metric due to this property in \S~\ref{sec:semaphor}, and show how it can be compensated with \textit{intention metrics}.

Another intuitive way to evaluate the agent consists of noticing how a policy $\pi^{PG}(\realstate{s})$ can be built by sampling from $P(a|disc(\realstate{s}))$, thus allowing to create agent surrogates. 
If the \gls{pg} creator has access to the environment for testing agent, they may compare the performance between them, as done in the literature~\citep{cortes_testing_2022,domenech_i_vila_explaining_2024}, by computing the difference in expected reward between the two policies (in episodes of length $T$)\footnote{This can be trivially extended to cases where performance is held out to the end of the episode as $\Delta R = \mathbb{E}[R(\pi)] - \mathbb{E}[R(\hat\pi)]$.}:

\begin{equation}\label{Eq:E_dif_reward}
    \Delta R(T)= \mathbb{E}[\sum_{t=1}^T R(s_t, \pi(s_t), s_{t+1})]  - \mathbb{E}[\sum_{t=1}^T R(s_t, \pi^{PG}(s_t), s_{t+1})]
\end{equation}

The intuition behind this metric is that the relevant predicates for explaining the agent's actions are also relevant for taking action. 
As such, the reward decay obtained by simplifying the agent can be linked to the decay in the reliability of our explanations. 
A decay close to zero implies both the original and the surrogate agents achieve similar performance. 
However, the fact that sometimes this value can be negative (\ie \gls{pg} agent obtains better rewards on average than the original agent) could mean that the \gls{pg} and original agent capture different policies, even when this metric is high. Some examples of this can be seen in \S~\ref{sec:discretisers}. 
Although potentially desirable from the performance side, this shines a doubtful light on whether the \gls{pg} gives reliable explanations over the agent, as it has captured something different.

\subsubsection{Intention Metrics}\label{sec:intention_metrics}

To gauge the explainability of the \gls{pg} with intentions, one should consider two things: how likely is it that $s$ (the state analysed) can be said to hold an intention $I_d$, and thus $s$ can be used to explain, and how likely is it that, if the \gls{pg} claims an intention for a state, that such intention holds?

As proposed in \S~\ref{sec:desiresintentions}, intentions should only be attributed to a state once they exceed a certain threshold: the commitment threshold $C>0$. 
This is because even though the agent may have some non-zero probability of achieving a desire in a state, an explanation claiming that the agent has such an intention is not desirable if the probability is very low, and thus defining a cutoff is important to avoid human bias.  
We define the set $S(I_d)=\{s\in S| I_d(s)>C\}$ as the set of states where the agent is attributed as having the intention $I_d$. 
In addition, we also consider the set $S(I)=\{s\in S| \exists d \in D: I_d(s)>C\}$, that is, the set in which the agent is attributed as having any of the considered desires as its intention.

Thanks to the classification of states into \textit{having} and \textit{not having} an intention, the probabilities used to evaluate desires in \S~\ref{sec:desires} can be trivially extended to answer the questions above:

\begin{enumerate}
    \item \textbf{Intention probability} $P(s\in S(I_d))$ is the probability that, at any point of observation, the agent is in a state $s$ which fulfills $I_d(s)>C$.
    \item \textbf{Expected Intention} $\mathbb{E}_{s\in S(I_d)}(I_d(s))$ is the probability that, once attributed, an intention is going to be fulfilled. It is computed as 
    
    $\mathbb{E}_{s\in S(I_d)}(I_d(s))=\sum_{s\in S(I_d)} I_d(s)*P(s)/P(s\in S(I_d))$.
\end{enumerate}

The first metric estimates the interpretability of agent behaviour: 
the less likely it is that the agent has no attributed intention in the state, the fewer times we will have no answer to why it is acting. The lower the commitment threshold, the larger the intention probability. For the case of $S(I_d)$, this score can also be increased by introducing more desires to check.

The second metric is an estimation of the reliability of an explanation. It computes likely is it that an explanation of \textit{why\/} it did something (the cause) did not result in it (the consequent) being fulfilled. 

Although both of these scores can reach a perfect state (value of $1$), real scenarios leave this option likely out of reach. 
On one hand, for a sufficiently low $C$ and enough desires considered, it is likely possible to reach perfect \textit{intention probability} (\ie always being able to attribute \textit{why}), but at the cost of being wrong several times. 
On the other hand, even with a high $C$ value, it is likely that an agent that has an intention to achieve something may fail due to unexpected environment changes.

\subsubsection{Are static metrics not enough?}\label{sec:semaphor}

\begin{figure}[H]
    \centering
\begin{subfigure}[b]{\textwidth}
	\centering
	\includegraphics[width=0.3\textwidth]{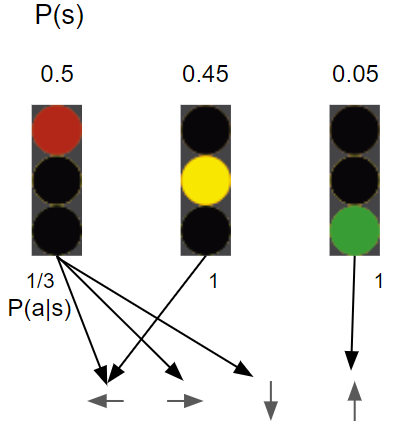}
	\caption{Environment with state probabilities and agent's policy}
	\label{fig:semaphor_environment}
\end{subfigure}
\begin{subfigure}[b]{0.45\textwidth}
	\centering
	\includegraphics[width=0.6\textwidth]{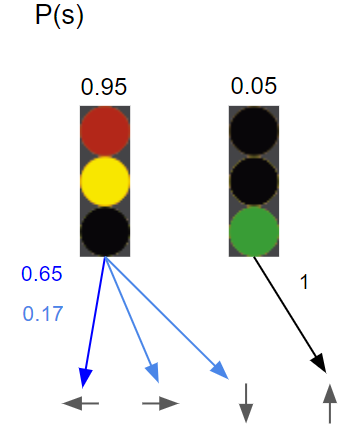}
	\caption{Environment-agent \gls{pg} modelled with smart discretiser}
	\label{fig:semaphor_smart}
\end{subfigure}
\begin{subfigure}[b]{0.45\textwidth}
	\centering
	\includegraphics[width=0.6\textwidth]{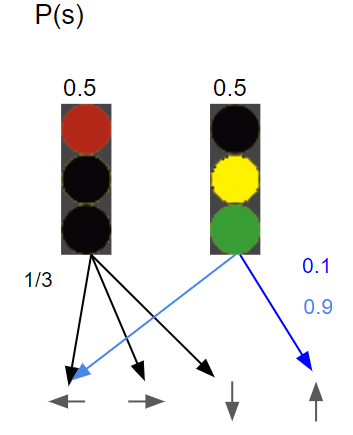}
	\caption{Environment-agent \gls{pg} modelled with a lousy discretiser}
	\label{fig:semaphor_dumb}
\end{subfigure}
    \caption{The semaphor environment and the proposed discretisers. Colours have been placed to distinguish between different state-action values of the agent’s policy when needed. The environment does not reward to go up when the red light is on, but rather to go up when the green light is on instead. This reward is more effectively represented in the smart discretiser than in the lousy discretiser, as the latter grants a probability to go up in the state with a green light that is lower than 100\%.}
    \label{fig:semaphor_plot}
\end{figure}

The metrics described in \S~\ref{sec:static_metrics}, as well as the ones used in the literature~\citep{liu_novel_2023}, have a considerable weakness, as they assume that the uncertainty of choosing a specific action has a comparable impact, regardless of the state, to the agent's behaviour.
However, in most real-world scenarios, it is seldom the case that behavioural certainty is critical, which means that, in most states, the action can be liberally chosen.
In these cases, the entropy metrics defined in \S~\ref{sec:static_metrics} need to be revised, given our lack of context on the criticality of the states. 
To illustrate this point, we present the \textit{traffic light} environment and agent, as shown in Figure~\ref{fig:semaphor_plot}.

Suppose an environment with three (undiscretised) states, the traffic light is \textit{R}ed, \textit{Y}ellow, \textit{G}reen, in which the agent can take four actions (going up, left, down, or right). 
The only rewarded transitions are going up on \textit{G}, which gives a positive reward, and going up on \textit{R}, which gives a negative reward. The next state after any action is not affected by the chosen action, to simplify computation. Instead, it has some strong bias toward \textit{Y} or \textit{R} (\eg 50 and 45\%), and a very low probability of going to \textit{G} (\eg 5\%).

Suppose now an agent that interacts with this environment as follows: in the \textit{R} state, it uniformly samples the action between left, down and right (33\%); in \textit{Y}, it always goes left, and in \textit{G} it always goes up. This agent has an optimal policy for this environment. Figure~\ref{fig:semaphor_environment} shows this arrangement.

Suppose now two \gls{pg}s, one which distinguishes the \textit{G} state from \textit{Y} and \textit{R} and one which does the same for \textit{R}. 
Neither will be a perfect surrogate, as all three states had a different probability distribution over actions. 
The first one can represent an optimal policy, whereas the second cannot (it does not distinguish \textit{Y} from \textit{G}, and the latter is reward-relevant).

However, when computing the agent entropy between these agents, it becomes apparent that $H_a$ is more significant (less desirable) for the first case ($0.89$) than for the second ($0.71$). This happens because the relevance of the critical case gets subsumed when considering large numbers. 
The probability of this happening increases with the size and complexity of the environment, but as it has been shown, it can happen in toy examples. 
Although removing from the entropy the weighting of states by their probability (heavily biasing toward infrequent states) or computing entropy on a subset of states found heuristically can reduce the relevance of the problem~\citep{liu_novel_2023}, it can become unreliable if the heuristic is miss-matched for the problem. 
For example, when choosing critical states where $H_a$ is low, the \gls{pg} could miss-report explainability in critical states where action entropy is large, such as the red traffic light in this example. 

Instead, modelling the environment such that critical decisions are accounted for to allow for the modelling of desires and intentions resolves this problem. Neither discretiser above would be able to model both the desire to go up in green and the desire to not go up in red directly (as there is the uncertainty of the state of the traffic light because of the predicates chosen). 
If the desires are formalised colourfully (\eg the desire not to go up is in any state with possibility of being red), then the intention metrics would report higher explainability in the smarter modelling.

\subsection{Revision pipeline}\label{sec:revision}

All previous metrics offer empirical, quantitative qualifiers of the designed \gls{pg} and can be used to report expected performance (both from the side of reliability and interpretability). 
However, the quality of the metrics and the explainability extracted depend highly on the \gls{pg} design, which is done with little information to start with. For this, we propose the revision pipeline.

To improve a \gls{pg}, agent trajectories  can be analysed through the intention function to gather \textit{why} the representation may be inaccurate, and enhance it. These trajectories may be actual agent observations or can be simulated by sampling the \gls{pg} if the agent cannot take new observations.

There are two prominent cases which can be detected and used to improve the graph:

\begin{itemize}
    \item \textbf{Unintentional regions}, or large sequence sections where no intention is manifested above the commitment threshold. There are two possibilities: either no agent desire exists, which can be manifested, or the explainee or the pipeline designer never declared the current desires. In such cases, the desire never gets registered in the \gls{pg} and is thus \textit{hidden} from the intention function.
    \item \textbf{Unfulfilled regions}, or sections of the sequence in which an agent had a behaviour from which the pipeline could infer the existence of an intention that is not fulfilled (due to the intention falling below the commitment threshold or despite a very high likelihood it does not get fulfilled for a long time). Causes for this could be the prioritisation of a different and conflicting intention, irrational agent behaviour, hidden desires, or the discretiser function confounds two different (real) states that do not manifest the same intention.
   
\end{itemize}

The agent's uncertainty is concisely presented by isolating these regions of interest. A human counterpart can analyse the regions and develop new hypotheses, such as new desires that could attribute intentional behaviour to the region, a priority ordering of desires, or improvable behaviours (\eg locating suboptimal policies which can be controlled or improved in the original policy).

\section{Experiments}\label{sec:experiments} 

So far, in this paper we have introduced the following contributions:

\begin{itemize}
    \item A methodology for producing explanations for agents' behaviour, based on constructing policy graphs from the agents' observation and discretising the state space and a set of desires (\S~\ref{sec:methodology}).
    \item Static metrics for analysing the structure of the policy graphs (\S~\ref{sec:static_metrics}).
    \item Intention metrics, capable of measuring both the interpretability of the agents' behaviour and the reliability of the explanations produced -- both in terms of attributable intentions derived from the proposed desires (\S~\ref{sec:intention_metrics}).
    \item A pipeline for interactive revision of the policy graphs by automatically identifying unintentional and unfulfilled regions of the timeline of the agents' behaviour (\S~\ref{sec:revision}).
\end{itemize}

In this section, we present empirical results for the application of these metrics and of the revision pipeline to a concrete use case: the Overcooked-AI environment~\citep{carroll_utility_2020}.

The experimentation methodology can be summarised as follows:

\begin{enumerate}
    \item We select some training methods and, for each layout, we train specialised agents from scratch.
    \item We analyse the performance of the resulting agents.
    \item We design a set of different discretisers that will allow us to compare the effect on the metrics of expressing the state with or without some specific predicates, and we propose a set of desires that are relevant for the Overcooked-AI scenario.
    \item We apply and analyse the static and intention metrics to the resulting policy graphs.
    \item We analyse the results of applying the revision pipeline to this environment, and we discuss the potential usage of this pipeline from a user perspective.
\end{enumerate}

All experiments have been done on the Overcooked-AI environment\footnote{\url{https://github.com/HumanCompatibleAI/overcooked\_ai}} introduced in \S~\ref{sec:use_case}, and the training code has been developed using Pantheon-RL\footnote{\url{https://github.com/Stanford-ILIAD/PantheonRL}}. The library for producing the policy graphs is \emph{pgeon}\footnote{\url{https://github.com/HPAI-BSC/pgeon}}, being developed by the authors, among other contributors. 
The policy graphs have been generated from observing 1500 episodes, with up to 400 steps per episode. The performance metrics have been computed as means and standard deviations of 500 episodes in random environments per agent. The hardware used was an Intel i7-5820k system with 96Gb of RAM and an Nvidia RTX 3090 GPU.

This experimentation section is structured as follows:
the choice of the training method for each agent is presented and motivated in \S~\ref{sec:agents}. 
The options for the discretisation of the state space and the static metric analysis are developed in \S~\ref{sec:discretisers}.
Finally, intention metrics are used to analyse each of the combinations in \S~\ref{sec:experimental_int_metrics}, and a case study is done with one of these and the revision pipeline to show the kind of explainability that can be produced in \S~\ref{sec:exp_revision}.

\subsection{Agents used}\label{sec:agents}

The agents analysed in this paper consist in two pairs of agents which collaborate with each other.

\begin{itemize}
    \item \textit{Pair A} (\marcone (Blue), \marctwo (Green)): two agents trained from scratch with \gls{ppo}\citep{schulman_proximal_2017}. These agents were used to validate \gls{pg}s in previous work~\citep{cortes_testing_2022}.
    \item \textit{Pair B} (\cmashuman (Green), \cmasone (Blue)): A human agent trained from human trajectories exclusively, and a \gls{ppo} agent trained to collaborate with it. These agents were used in previous work~\citep{carroll_utility_2020,tormos_llorente_explainable_2023}. 
    It is important to remark that some behaviours learnt by the \gls{ppo} agent trained to collaborate with the human are suboptimal given the lack of co-adaption. For example, through experimental results shown in Figure~\ref{fig:desire_cmas_unidents_11}, we verify that for the {\unident} layout, the behaviour of the {\cmasone} is random and did not train correctly despite its apparently high-performance metrics in Table~\ref{sec:static_metrics}.
    \item \textit{Random baseline} (\randomagent (Blue), \marctwo (Green)): same as \textit{Pair A} but {\marcone} is substituted by an agent that samples actions from a uniform probability distribution (all actions have probability $20\%$ regardless of the state). This agent is used as a baseline for comparison with the other two pairs.
\end{itemize}

For each unique layout, the agents were trained from scratch, so there are a total of twenty different agents.

\begin{table}[htb]
    \centering
    \small
    \begin{tabular}{c | p{0.255\linewidth} | p{0.255\linewidth} | p{0.255\linewidth}}
         & \marcone, \marctwo & \cmashuman, \cmasone & \randomagent, \marctwo \\
        \hline
        \textit{simple} & 387.87 (25.33) & 251.26(31.62) & 21.55 (16.71) \\
        \textit{random1} & 266.01 (48.11) & 187.19 (28.53) & 36.70 (11.48)\\
        \textit{random3} & 62.5 (5.00) & 81.93 (21.79) & 0.53 (1.47)\\
        \textit{unident\_s} & 757.71 (53.03) & 102.12 (28.11) & 4.30 (7.30)\\
        \textit{random0} & 395.01 (54.43) & 107.99 (46.45) & 7.61 (6.03)\\
    \end{tabular}
    \caption{Performance evaluation (mean and standard deviation) of the trained agent pairs. For the case of {\unident}, the human-agent pair obtains results only due to the human agent doing all the work.}
    \label{tab:agent_metrics}
\end{table}

\subsection{Discretisers and Static metrics}\label{sec:discretisers}

Four discretisers are tried and tested for each of the agents and environments. From 1 to 4, each is more expressive and increases complexity (and entropy). 
The main discretiser includes all predicates relevant to behaving in the environment, including state of the pots and relative positions of objects (which drastically reduce complexity). 
Each of the extensions is focused on increasing information on the other agent's state. Table~\ref{tab:discretisers} gives the full description of each discretiser, and Table~\ref{tab:reported_static_metrics} illustrates the static metrics for a subset of agents and layouts.

\begin{table}[t!]
    \centering
    \caption{Variables used to describe the domain by each discretiser. Predicate computation is done via the environments' \textit{MediumLevelPlanner}. Each variable may take only one value in a state. \texttt{held} and \texttt{held\_partner} represent the object the agents are holding, where \texttt{O,T,D,S} stand for the items that can be held (onion, tomato, dish, soup). \texttt{item\_pos} shows the optimal next action to get to a certain item (be it an item source or not), where \texttt{U,D,L,R,I,S} for the actions to reach an item (go up, down, left, right, interact or stay). \texttt{partner\_zone} refers to the cardinal direction (\texttt{N,NE...}) in which the other agent is located with respect to the PG agent. Note that \texttt{N,W,S,E} are only used when the two agents are in the same horizontal or vertical axis.}
    \label{tab:discretisers}
    \begin{tabular}{|c|c|}
    \hline
         & Variables (domain) \\
        \hline
        \multirow{3}{*}{D1} & 
    $ \texttt{held(O, T, D, S, $\varnothing$)}$\\
    & $\texttt{pot\_state(Empty, Waiting, Cooking, Finished)}$\\
    & $\texttt{item\_pos(\texttt{U}, \texttt{D}, \texttt{L}, \texttt{R}, \texttt{I}, \texttt{S})}, \forall item \in \{\texttt{O}, \texttt{T}, \texttt{D}, \texttt{Pot}, \texttt{service}\} $\\
    \hline
        D2 & $D1 \cup \{\texttt{held\_partner(O, T, D, S, $\varnothing$)}\}$\\
        
    \hline
        D3 & $D1 \cup \{\texttt{partner\_zone(\texttt{N}, \texttt{NE}, \texttt{E}, \texttt{SE}, \texttt{S}, \texttt{SW}, \texttt{W}, \texttt{NW})}\}$\\
    \hline
        D4 & $D2 \cup D3$
        \\\hline
    \end{tabular}
\end{table}

\begin{table}[!htbp]
    \centering
    \footnotesize
    \begin{tabular}{ccc|cccc}
    Layout & Agent & D & $H$ & $H_a$ & $H_w$ & Mean $\Delta R$ \\
    \multirow{12}*{\simple} &  \multirow{4}*{\cmasone} & 
        1 & \textbf{1.98} & 1.46 & \textbf{0.52} & -60.96 \\
    & & 2 & 2.15 & 1.41 & 0.74 & -34.66 \\
    & & 3 & 2.10 & 1.38 & 0.72 & -25.26 \\
    & & 4 & 2.21 & \textbf{1.31} & 0.90 & \textbf{-7.36} \\
    \cline{2-7}
    &  \multirow{4}*{\marcone} & 
        1 & \textbf{2.13} & 1.68 & \textbf{0.44} & -19.39 \\
    & & 2 & 2.40 & 1.62 & 0.78 & -15.51 \\
    & & 3 & 2.47 & 1.50 & 0.98 & -7.76 \\
    & & 4 & 2.45 & \textbf{1.43} & 1.02 & \textbf{-3.88}\\
    \cline{2-7}
    &  \multirow{4}*{\randomagent} & 
        1 & \textbf{3.37} & 2.57 & \textbf{0.80} & 0.69 \\
    & & 2 & \textbf{3.39} & 2.56 & \textbf{0.83} & -0.17 \\
    & & 3 & 3.60 & 2.56 & 1.05 & -0.05 \\
    & & 4 & 3.56 & \textbf{2.54} & 1.02 & \textbf{0.98} \\
    \hline

    \multirow{12}*{\randomzero} &  \multirow{4}*{\cmasone} & 
        1 & \textbf{2.17} & 1.70 & \textbf{0.48} & -107.99 \\
    & & 2 & 2.25 & 1.57 & 0.68 & -107.99 \\
    & & 3 & 2.44 & 1.65 & 0.79 & 0.61 \\
    & & 4 & 2.40 & \textbf{1.49} & 0.91 & \textbf{8.61} \\
    \cline{2-7}
    &  \multirow{4}*{\marcone} & 
        1 & \textbf{1.54} & 1.03 & \textbf{0.50} & -19.75 \\
    & & 2 & 1.60 & 0.98 & 0.62 & -15.80 \\
    & & 3 & 1.65 & 0.98 & 0.67 & \textbf{-11.85} \\
    & & 4 & 1.68 &\textbf{ 0.93} & 0.75 & -19.75 \\
    \cline{2-7}
    &  \multirow{4}*{\randomagent} & 
        1 & \textbf{2.96} & 2.58 & \textbf{0.38} & -0.23 \\
    & & 2 & 2.97 & \textbf{2.57} & 0.40 & \textbf{-0.04} \\
    & & 3 & 2.97 & \textbf{2.57} & 0.39 & -0.76 \\
    & & 4 & 2.97 & \textbf{2.57} & 0.40 & -0.07\\
    \hline

    \multirow{12}*{\unident} &  \multirow{4}*{\cmasone} & 
        1 & \textbf{2.14} & 1.86 & \textbf{0.27} & -13.02 \\
    & & 2 & 2.26 & \textbf{1.76} & 0.49 & \textbf{-10.82} \\
    & & 3 & 2.47 & 1.85 & 0.62 & -13.22 \\
    & & 4 & 2.49 & \textbf{1.74} & 0.76 & -13.72 \\
    \cline{2-7}
    &  \multirow{4}*{\marcone} & 
        1 & \textbf{1.37} & 0.90 & \textbf{0.47} & -7.58 \\
    & & 2 & 1.65 & 0.88 & 0.77 & -7.58 \\
    & & 3 & 1.82 & 0.86 & 0.96 & -7.58 \\
    & & 4 & 1.89 & \textbf{0.84} & 1.06 & -7.58 \\
    \cline{2-7}
    &  \multirow{4}*{\randomagent} & 
        1 & \textbf{3.15} & 2.58 & \textbf{0.57} & -0.10 \\
    & & 2 & \textbf{3.16} & 2.58 & \textbf{0.58} & -0.23 \\
    & & 3 & 3.56 & \textbf{2.57} & 0.98 & -0.05 \\
    & & 4 & 3.52 & \textbf{2.57} & 1.96 & \textbf{0.57} \\
    \end{tabular}
    \caption{Static metrics for a subset of agents analysed. The best metric per agent and layout is marked in bold casing. Note how $H_w$ always increases with complexity of the discretiser, whereas $H_a$ does not always decrease (especially in bad-performing agents). Although there exists a correlation between $H_a$ and $\Delta R$, results are inconclusive given the variability of $\Delta R$. {\randomagent} (the baseline) shows that a policy independent on the predicates introduced cannot reduce the \gls{pg}'s $H_a$.}
    \label{tab:reported_static_metrics}
\end{table}

The results indicate a complex trade-off between the reliability and interpretability of the \gls{pg}s. There is no clear winner in all categories. Still, ultimately,  the representations with a richer -- and therefore more complex -- set of predicates represent the agent's behaviour more faithfully in the general case (as illustrated by the mean $\Delta R$). 
Larger graphs mean more information for the agent's actions, but as can be seen from {\cmasone} in \unident, if the agent is not well-performing (or ignores the added information), $H_a$ may not decrease enough to justify the drastic increase in $H_w$. 
In the case of a tie, $\Delta R$ can be a reasonable estimate of whether the \gls{pg} correctly captures agent behaviour, and thus, explainability extracted from it is reliable.

\subsection{Intention metrics}\label{sec:experimental_int_metrics}

Static metrics offer direct, unbiased insight over the \gls{pg}s structurally. 
When the differences are significant enough, agents can use them to tell which families of discrete options trump the rest reliably. 
However, the relationship between static metrics and \gls{pg} adequacy is challenging to understand. When the difference in metrics between the two options is too small, it becomes easier to evaluate the methods from the optic of the maxims of communication or the correctness of explanations that the \gls{pg} may produce.

To better evaluate the quality of explanations, it becomes necessary to hold insights into the agent's goals and objectives, which, in the case of this paper, requires external (human) information. 
In \S~\ref{sec:desiresintentions}, a formalisation of desires is introduced, allowing the \gls{pg} to manifest beliefs over beneficial agent behaviour. 
By extending desires into the past, it becomes possible to evaluate what possible beneficial behaviour the agent is likely to manifest in the future (\ie what intentions it holds). However, external insights into the agents' goals may be biased or outright wrong. As such, it becomes necessary to evaluate the adequacy of the \gls{pg} and the human-hypothesised agent's desires.

In exchange for this added complexity, it becomes possible to directly evaluate the trade-off between the reliability and interpretability of the agent's behaviour. The formal definition for measuring these can be found in \S~\ref{sec:intention_metrics}. 

Given a \gls{pg} and a desire $d$, informally, the reliability $\mathcal{R}$ of the explanations generated using a \gls{pg} regarding that desire is equal to the probability that, once an intention corresponding to that desire is attributed to a state, this intention will be fulfilled. This can be easily rewritten to talk about `any desire' by taking the $max_d I_d(s_t)$ within the expectancy.

\begin{equation}\label{Eq:reliability_desire}
    \mathcal{R}_d(T) = \mathbb{E}_{s\in S_d}(I_d(s_t)) = \frac{\sum_{s \in S_d} P(s)*I_d(s_t)}{\sum_{s \in S_d} P(s)}
\end{equation}

The interpretability $\mathcal{I}$ of behaviour over a desire $d$ is defined as the proportion of time in which the agent is found in a state where it is attributed to having an intention to do $d$ (\ie the state probability):

\begin{equation}\label{Eq:interpretability_desire}
    \mathcal{I}_d = \mathbb{E}_{s\in S}([s\in S_d]) = \sum_{s \in S_d} P(s)
\end{equation}

where $[s\in S_d]$ is the Iverson bracket.

\begin{figure}[H]
    \centering
\begin{subfigure}[b]{0.4\textwidth}
	\centering
	\includegraphics[width=\textwidth]{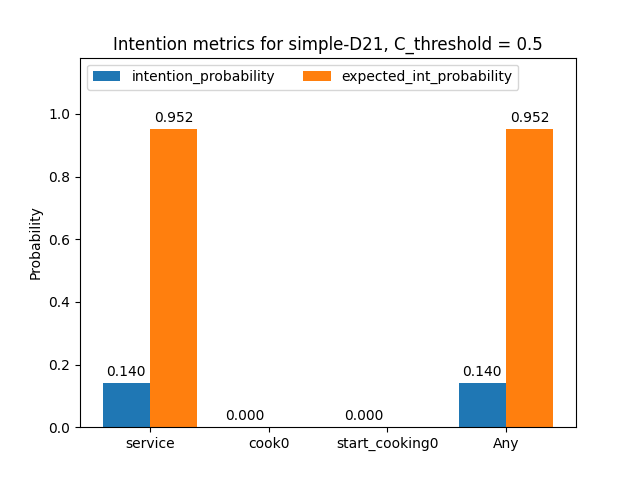}
	\label{fig:intention_marcone_simple_21}
\end{subfigure}
\begin{subfigure}[b]{0.4\textwidth}
	\centering
	\includegraphics[width=\textwidth]{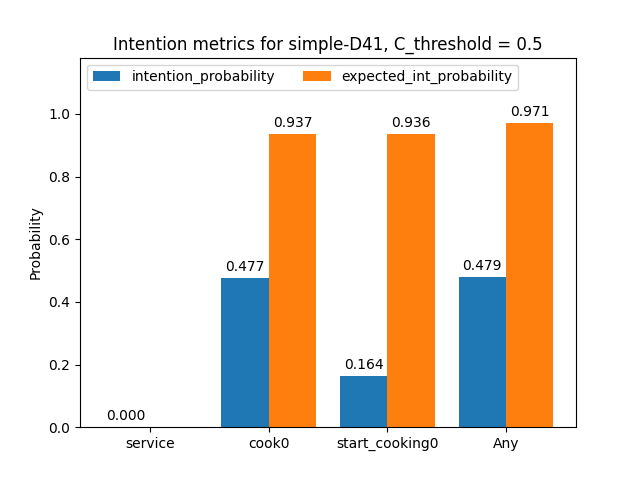}
	\label{fig:intention_marctwo_simple_41}
\end{subfigure}
\begin{subfigure}[b]{0.4\textwidth}
	\centering
	\includegraphics[width=\textwidth]{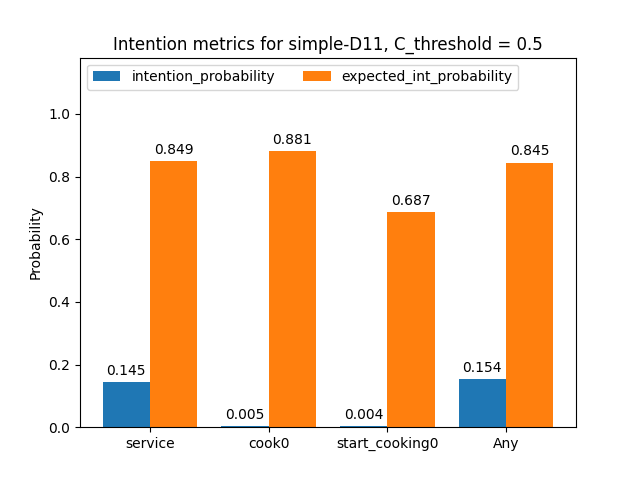}
	\label{fig:intention_cmasone_simple_11}
\end{subfigure}
\begin{subfigure}[b]{0.4\textwidth}
	\centering
	\includegraphics[width=\textwidth]{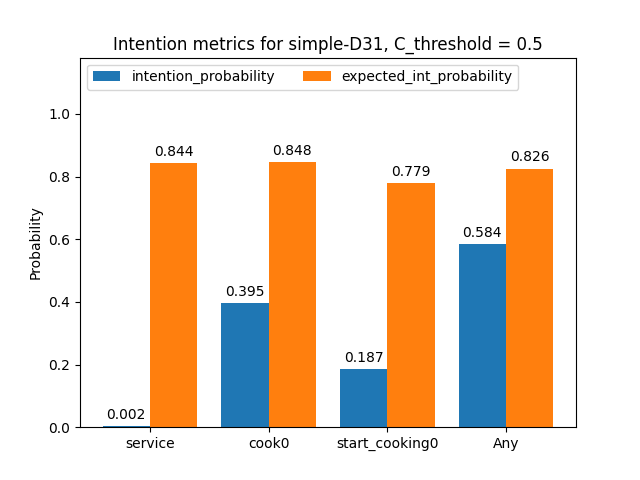}
	\label{fig:intention_cmashuman_simple_31}
\end{subfigure}
\begin{subfigure}[b]{0.4\textwidth}
	\centering
	\includegraphics[width=\textwidth]{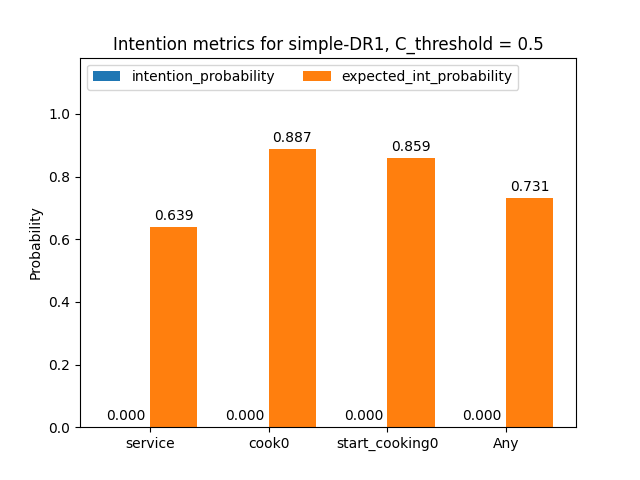}
	\label{fig:intention_random_simple_R1}
\end{subfigure}
    \caption{Intention metrics for Layout {\simple} for each of the 4 agents (in order, \marcone, \marctwo, \cmasone, \cmashuman, and {\randomagent}) using discretiser 1.  Collaboration and specialisation can be seen (of each pair, one agent specialises in serving and another in cooking). Both {\marcone} and {\cmasone} agents specialise on delivering soup, and conversely {\marctwo} and {\cmashuman} work on cooking. With a $0.5$ commitment threshold, expected intention fulfillment is very high for all cases, but overall agent interpretability is low ($15\%$ of the time) for agents specialising in delivering soup (as they spend most of the time apparently idle). {\randomagent} shows apparently high reliability in fulfilling intentions: this corresponds to states in which executing random actions eventually results in fulfilling a desire. These states happen with a probability $<0.1\%$.}
    \label{fig:intention_metrics_simple}
\end{figure}

\begin{figure}[H]
    \centering
\begin{subfigure}[b]{0.4\textwidth}
	\centering
	\includegraphics[width=\textwidth]{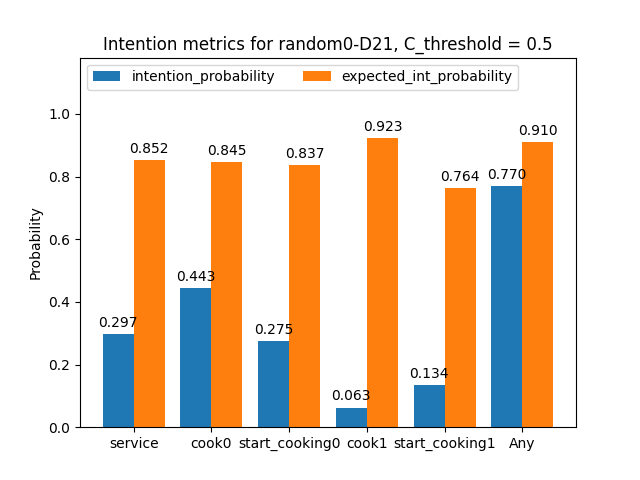}
	\label{fig:intention_marcone_random0_21}
\end{subfigure}
\begin{subfigure}[b]{0.4\textwidth}
	\centering
	\includegraphics[width=\textwidth]{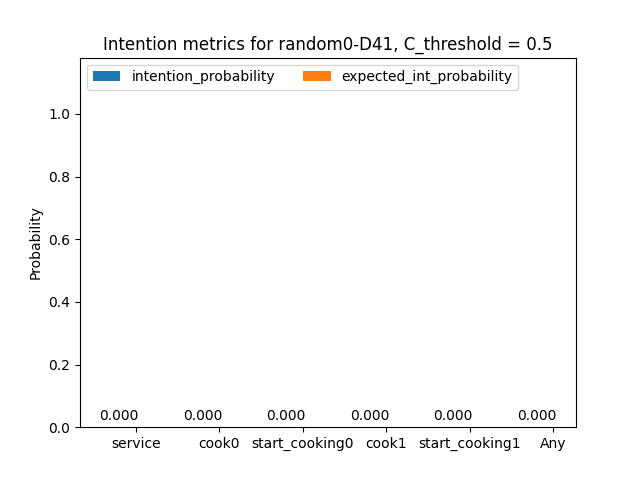}
	\label{fig:intention_marctwo_random0_41}
\end{subfigure}
\begin{subfigure}[b]{0.4\textwidth}
	\centering
	\includegraphics[width=\textwidth]{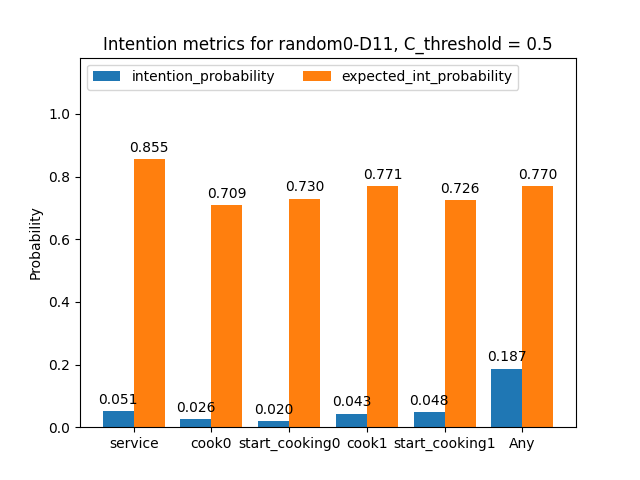}
	\label{fig:intention_cmasone_random0_11}
\end{subfigure}
\begin{subfigure}[b]{0.4\textwidth}
	\centering
	\includegraphics[width=\textwidth]{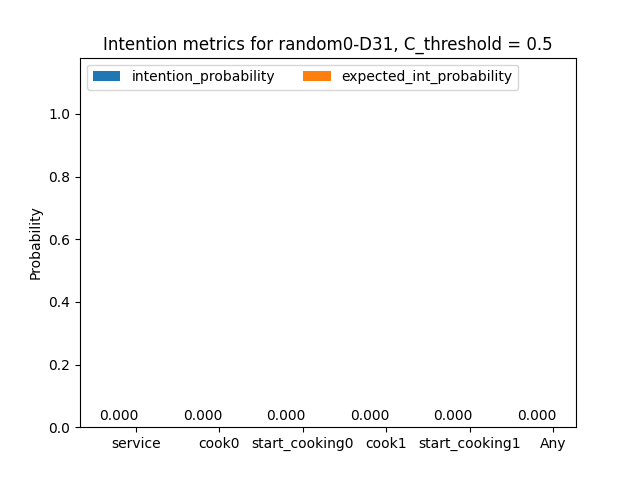}
	\label{fig:intention_cmashuman_random0_31}
\end{subfigure}

\begin{subfigure}[b]{0.4\textwidth}
	\centering
	\includegraphics[width=\textwidth]{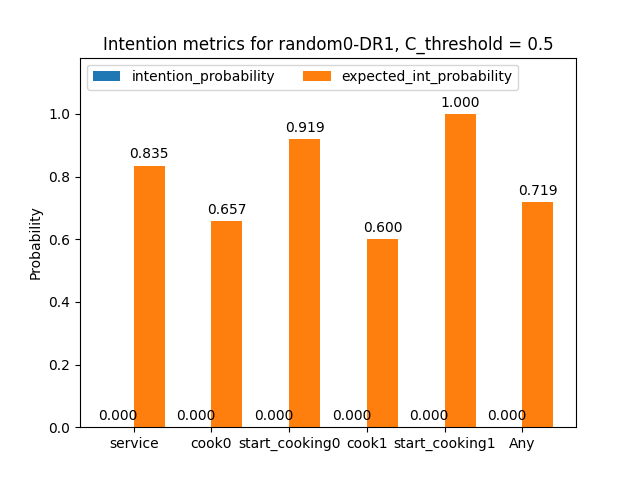}
	\label{fig:intention_random_random0_R1}
\end{subfigure}
    \caption{Intention metrics for Layout {\randomzero} for each of the 4 agents (left to right, \marcone, \marctwo, \cmasone, and \cmashuman) using discretiser 1. Intention probability (tied to interpretability) is in blue, and Expected Intention Probability is shown in orange). With a $0.5$ commitment threshold, {\marcone} has remarkably high metrics: $77\%$ of the time there is an attributed intention which gets fulfilled with $91\%$ certainty. The lack of access to the pot and service zone for {\marctwo} and {\cmashuman} means that their behaviour is not interpretable with these desires, and new ones should be considered (such as placing an onion or a plate on the counter). Much like before, {\randomagent} has high reliability. Given the constrained space of the layout, it may be easier to randomly fulfill desires, but again, the probability of manifesting intentions is low.}
    \label{fig:intention_metrics_random0}
\end{figure}

Figures~\ref{fig:intention_metrics_simple} and~\ref{fig:intention_metrics_random0} show these metrics for the four agents in the same layouts (\simple~and~\randomzero) and a single commit-threshold. This information can be used to gauge how likely the method is for providing satisfying explanations to the explained. 
Each desire can be analysed separately, and the hypothesised desires can be verified. If there is no commitment threshold in which the two metrics are decently high, it becomes apparent that the desires do not capture the agent's behaviour. This can be either because the agent did not train correctly (making the hypothesised desires something it cannot reach) or because the agent is targeting a different set of desires. 
This last case is apparent in Figure~\ref{fig:intention_metrics_random0}: the two empty boxes correspond to agents with no access to the pot or the service, and thus these desires never get fulfilled.

\begin{figure}[H]
    \centering
\begin{subfigure}[b]{0.325\textwidth}
	\centering
	\includegraphics[width=\textwidth]{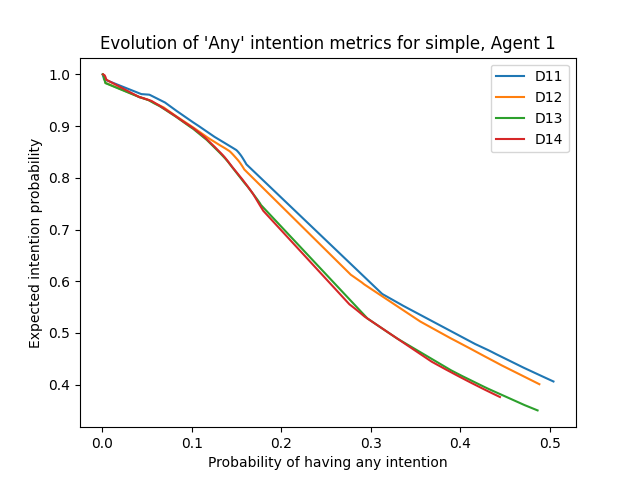}
	\label{fig:ROC_simple_cmas1}
\end{subfigure}
\begin{subfigure}[b]{0.325\textwidth}
	\centering
	\includegraphics[width=\textwidth]{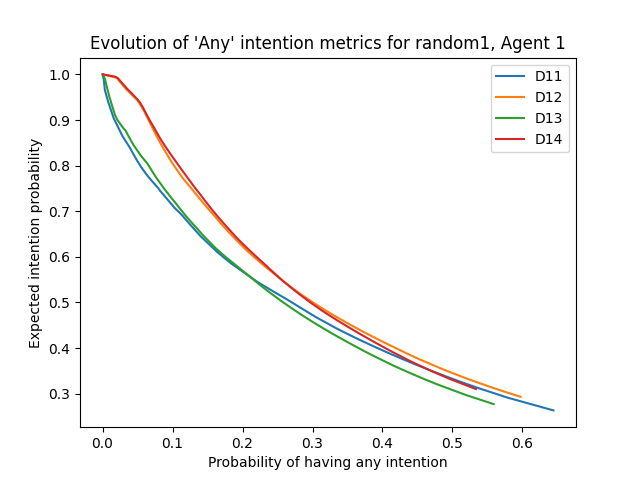}
	\label{fig:ROC_random1_cmas1}
\end{subfigure}
\begin{subfigure}[b]{0.325\textwidth}
	\centering
	\includegraphics[width=\textwidth]{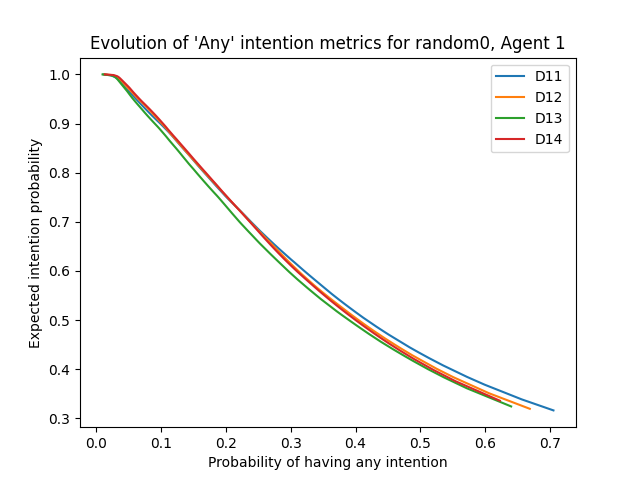}
	\label{fig:ROC_random0_cmas1}
\end{subfigure}
\begin{subfigure}[b]{0.325\textwidth}
	\centering
	\includegraphics[width=\textwidth]{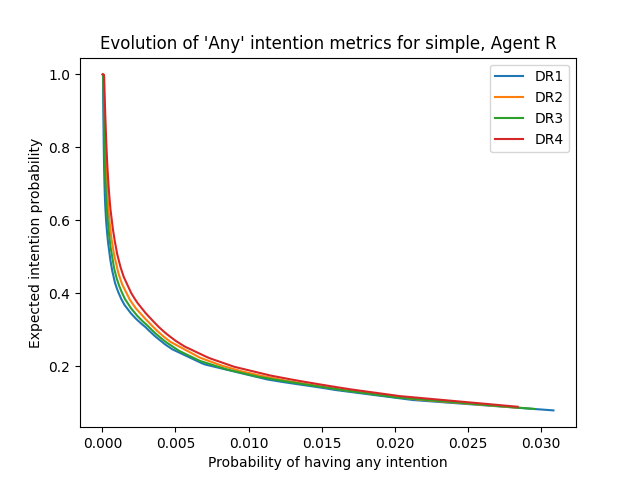}
	\label{fig:ROC_simple_R1}
\end{subfigure}
\begin{subfigure}[b]{0.325\textwidth}
	\centering
	\includegraphics[width=\textwidth]{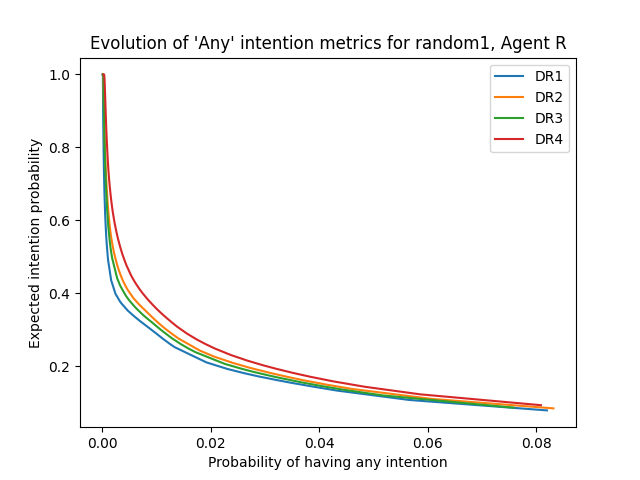}
	\label{fig:ROC_random1_R1}
\end{subfigure}
\begin{subfigure}[b]{0.325\textwidth}
	\centering
	\includegraphics[width=\textwidth]{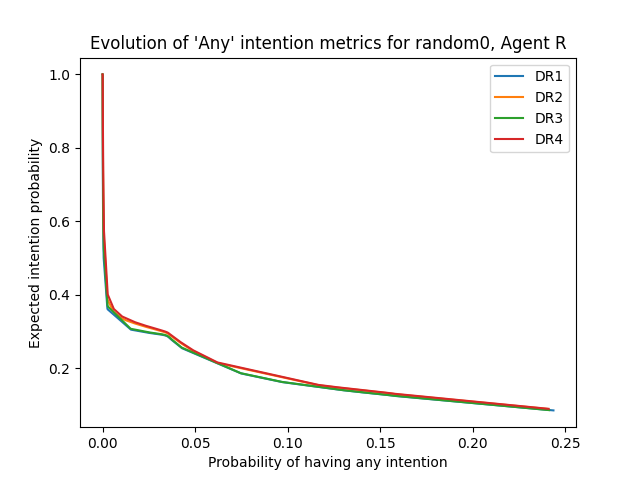}
	\label{fig:ROC_random0_R1}
\end{subfigure}
    \caption{Attributed intention probability (interpretability) and expected intention probability (reliability) progression as the commitment threshold changes, for all 4 discretisers and agent {\cmasone} (row 1) and {\randomagent} (row 2). For the {\simple} environment and {\cmasone}, we would prefer simpler discretisers (1 and 2), whereas for {\randomone}, it it seems important (especially with high commitment threshold) to know what the other agent is holding (\ie discretisers 2 and 4). The differences are minimal, as the discretisers vary in little number of predicates, but still noticeable. Contrary to what would be expected, from the $\Delta R$ in Table~\ref{tab:agent_metrics}, for {\randomzero} there is little difference in the metrics of the optimizer. {\randomagent} displays, for all environments, a very low area under the curbe }
    \label{fig:roc-curve}
\end{figure}

Analysing each of these metrics to pick the best discretiser and commitment threshold can be challenging. To simplify the process, a ROC\footnote{Receiver operating characteristic curve.}-like curve is proposed, plotting the interpretability against the reliability in Figure~\ref{fig:roc-curve}. 
In doing so, the fitness of each discretiser is displayed for each domain, and the designer can have a better pick of discretiser depending on the desired interpretability-reliability trade-off.

\subsection{Revision pipeline example}\label{sec:exp_revision}

The analysis of the intention metrics defined above can be used to verify that the agent behaves as desired (or as hypothesised). However, knowing what proportion of the graph (and thus, behaviour) is explainable is insufficient to bridge the gap and  discard unexplainable behaviour. 
Instead of manually inspecting all possible states in the graph in which the agent is attributed to having no intention, a reasonable alternative is to analyse the explanations provided across the timeline of the environment execution.
\begin{figure}[H]
    \centering
    \includegraphics[width=0.9\linewidth]{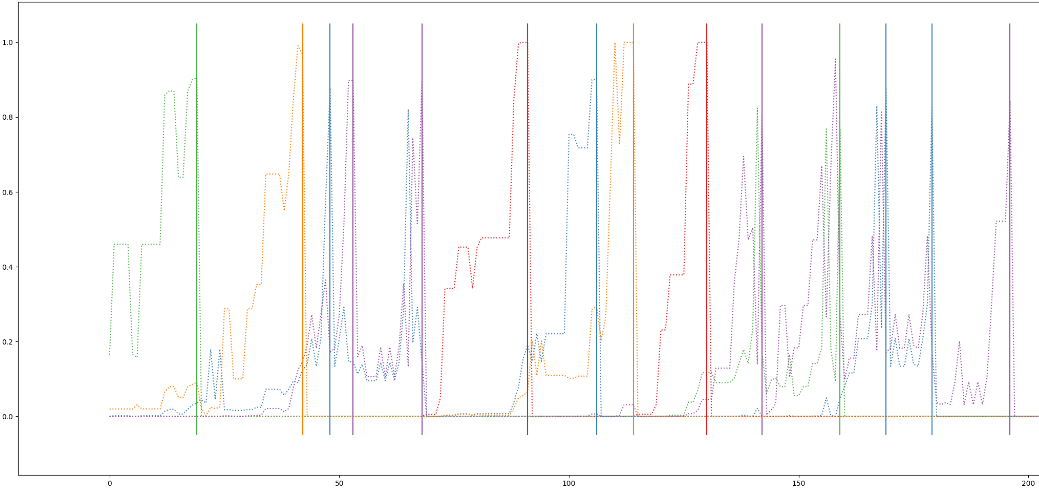}
    \includegraphics[width=0.9\linewidth]{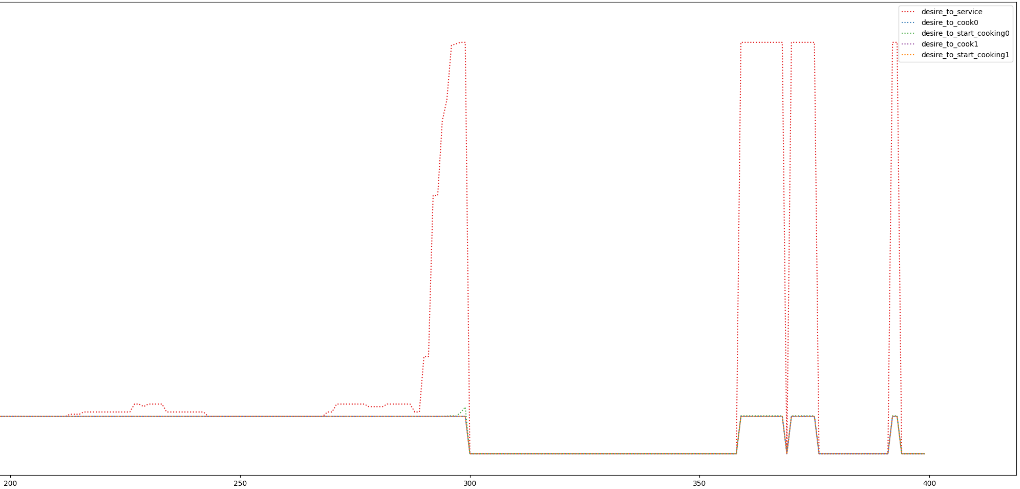}
    \caption{Revision pipeline run on {\cmasone} in environment {\randomzero}.
    Intention progression is marked with dotted lines, and desire completion with vertical solid lines. Regions with intention lower than 0 mark the agent is in an unseen state by the \gls{pg}. Each colour represents a desire: red for service, blue and purple for cooking, and green and orange for starting to cook (in each pot). Intentions that get high enough are consistently fulfilled so long as two contrary intentions coexist (\eg time-steps 40 to 70 where the agent has intention to cook in both pots (blue and purple) but finally decides to use Pot1 (purple). The region spanning 200 to 280 is revealed to be inexplicable by the algorithm, which prompted further analysis: in this region, the agent was blocked as {\cmashuman} was not passing a plate over the counter. Finally, in regions spanning 300 to the end, the agent behaves incoherently with the assumed intentions and reaches a state that was never seen prior and is not in the \gls{pg}: being in the lowermost tile by the service, with soup on hand, and having an onion set in front of him in the counter. The agent behaviour then alternates between interacting with the tile holding the onion and changing the direction it is facing.
    }
    \label{fig:random0revision}
\end{figure}

For the purpose of exemplifying this we choose an agent-layout pair: {\cmasone} and {\randomzero}. The original trained agent performs a run on the environment, recording all states (and corresponding discretised states). Finally, the states' intention progression through time is plotted, as was described in \S~\ref{sec:revision}.

Figure~\ref{fig:random0revision} shows one of the runs performed, which illustrates several insights that can be obtained from performing this revision,
and shows two unfulfilled regions and one unintentional region.

The first unfulfilled region is a case of prioritising cooking in pot$_1$ instead of pot$_0$. It is necessary to hold an onion to achieve the desires of \textit{cooking} or \textit{starting to cook}, 
The only difference between cooking in either pot is whether the agent goes up or right in the time step before interacting with the pot. In this case, the agent's usage of the pots could be more coherent and consistent, alternating the pots in no particular order. 
This behaviour, while currently unpredictable, holds potential for improvement. A deeper understanding of the algorithm makes it clear that these actions result from random decisions. The time delay between the final onion being placed in the pot and the soup being ready indicates room for optimisation. Prioritising a pot without cooking could be a simple yet effective solution. This adjustment could resolve the first unfulfilled region, showcasing the agent's potential for enhanced performance.

The second unfulfilled region presents a more intricate challenge, where the agent is on the brink of successfully serving soup but falls short. The agent appears to get stuck attempting to pick an onion from over the counter despite already holding soup. This complexity will pique the curiosity of researchers and developers, encouraging them to delve deeper into the agent's behaviour. 

We hypothesise the agent has learnt that keeping the counter between agents empty (particularly of onions) is the cornerstone to obtaining a reward, as plates cannot be passed over. It has never seen a situation where onions were over the counter while it held soup, thus triggering confusion. 
The agent's behaviour can be significantly enhanced by fine-tuning the agent for these specific cases, assuring researchers and developers of the agent's potential for improvement.

Finally, the unintentional region is easier to analyse. When checking the states in the unintentional region, we observe that both pots currently hold soup. 
This means that the only productive action the agent can do is deliver it, for which it needs a plate. However, the paired agent does not offer a plate for a long time, instead opting to put more onions on the counter. This behaviour is probably the trigger for the previous unfulfilled region, as the counters are very full of onions.

\section{Discussion and Future Work}\label{Discussion}

The framework proposed allows attributing intentions and extending \gls{pg} explanations into the teleological. The encoded information of desires (\S~\ref{sec:desiresintentions}) provides new types of explanations such as \textit{What do you intend to do now?}, \textit{How do you plan to do it?}, and \textit{For what purpose did you take this action now?} (\S~\ref{sec:explanation_algorithms}) in a concise and composable manner. In addition, the \gls{pg} model is instrumented with metrics (\S~\ref{sec:metrics}) to evaluate the 
reliability and interpretability of the behaviour and the trade-off is made explicit with the introduction of a user-defined parameter: the commitment threshold (\S~\ref{sec:intentions}). 

Although this process requires external knowledge and is not out-of-the-shelf, the provided heuristics (\S~\ref{sec:design_heur}), as well as the revision pipeline (\S~\ref{sec:revision}) enable guided iteration over the modelling by gathering and exposing its shortcomings naturally. We believe that the whole proposed methodology can be applied to many tasks (Figure~\ref{fig:graphical_abstract}).

As an outcome of this process, we are optimistic about using this method for applications besides human explainability.  
One of the key contributions of this paper is that, by using the method proposed, there is a way of automatically creating policies for easily understandable agents that mimic the behaviour of an original agent, thus enabling our method as a Theory of Mind model for understanding the behaviour of others in \gls{ma} systems.
In addition, the availability of intentions for states may be useful for better designing rewards for \gls{rl} agents (\eg by locating sparse regions and populating them to go toward near intention-attributed regions), or improving other types of agent implementations.
Finally, we believe the insights provided in this paper about the necessity of having a world-model (\ie  $P(s'|a,s)$) and how it enables teleological explanations will be key in designing transparent agents. The introduction of such models may also help the \gls{rl} community~\citep{touati_does_2023}.

\subsection{Limitations}

Looking forward, there are some improvements that can be applied to our proposed approach. Mainly, the construction of a \gls{pg} imposes additional requirements on the explainee:

\textit{Necessity of outer desires.} As part of the process, it is necessary for the explainee to provide formal descriptions of desires. 
When attempting to discover an agent's desires based on statistics alone (\eg through notions of criticality or low entropy), spurious correlations may result in providing nonsensical explanations or distorting the value of the method (\S~\ref{sec:semaphor}). 
Moreover, desirable actions discovered automatically burden the explainee with finding the reason why those are desires. When provided externally, the reasons for desirability are patent for the user (as they already believed the behaviour to be desirable) and thus they only need to be tested.

\textit{Limitations of state discretisation.} Finding a good state representation for \gls{pg}s to work is critical.  
Beyond computational and data requirements, the simplification is done so state descriptions are in a shared code between the explainee and the explainer. 
These descriptions are necessary when performing the original types of explanations~\citep{hayes_improving_2017} as well as the \textit{how} question in \S~\ref{sec:explanation_algorithms}.
However, finding how to discretise the environments can be challenging when considering complex environments (such as those with image input). Even with optimal automatic discretisation~\citep{silver_predicate_2023}, environments with large, complex state spaces such as chess will lose essential information to provide explanations. These environments remain as future work for \gls{pg}s. 

\printglossaries

\bibliography{main}

\end{document}